\documentclass[runningheads]{llncs}

\usepackage{eccv}

\usepackage{eccvabbrv}

\usepackage{graphicx}
\usepackage{booktabs}
\usepackage{threeparttable}

\usepackage[accsupp]{axessibility}  

\usepackage{wrapfig}
\usepackage{adjustbox}

\usepackage[pagebackref,breaklinks,colorlinks]{hyperref}

\usepackage{orcidlink}
\usepackage{multirow}

\begin{document}

\title{Towards Multimodal Open-Set Domain Generalization and Adaptation through Self-supervision} 


\titlerunning{Multimodal Open-set Domain Generalization and Adaptation}

\author{Hao Dong\inst{1}\orcidlink{0009-0000-8562-1946} \and
Eleni Chatzi\inst{1}\orcidlink{0000-0002-6870-240X} \and
Olga Fink\inst{2}\orcidlink{0000-0002-9546-1488}}

\authorrunning{H.~Dong et al.}

\institute{ETH Z\"urich\\
\email{\{hao.dong, chatzi\}@ibk.baug.ethz.ch} \and
EPFL\\
\email{olga.fink@epfl.ch}}

\maketitle

\begin{abstract}
The task of open-set domain generalization (OSDG) involves recognizing novel classes within unseen domains, which becomes more challenging with multiple modalities as input. Existing works have only addressed unimodal OSDG within the meta-learning framework, without considering multimodal scenarios.  
In this work, we introduce a novel approach to address Multimodal Open-Set Domain Generalization (MM-OSDG) for the first time, utilizing self-supervision. To this end, we introduce two innovative multimodal self-supervised pretext tasks: Masked Cross-modal Translation and Multimodal Jigsaw Puzzles. These tasks facilitate the learning of multimodal representative features, thereby enhancing generalization and open-class detection capabilities. 
Additionally, we propose a novel entropy weighting mechanism to balance the loss across different modalities.
Furthermore, we extend our approach to tackle also the Multimodal Open-Set Domain Adaptation (MM-OSDA) problem, especially in scenarios where unlabeled data from the target domain is available.  Extensive experiments conducted under MM-OSDG, MM-OSDA, and Multimodal Closed-Set DG settings on the EPIC-Kitchens and HAC datasets demonstrate the efficacy and versatility of the proposed approach. Our source code is publicly available\footnote[1]{\href{https://github.com/donghao51/MOOSA}{https://github.com/donghao51/MOOSA}}. 
  \keywords{Multimodal Learning \and Open-Set Domain Generalization \and Open-Set Domain Adaptation}
\end{abstract}

\section{Introduction}
\label{sec:intro}
Domain generalization (DG) and adaptation (DA) significantly improve the robustness and adaptability of machine learning models, enabling them to perform effectively across diverse and previously unseen environments~\cite{9782500}. This enhancement ensures that the models are more readily transferable to real-world applications, such as autonomous driving~\cite{SuperFusion,dong2023jras} and action recognition~\cite{Damen2018EPICKITCHENS}. To address  the challenges posed by distribution shifts, a wide range of DG and DA algorithms have been introduced. These algorithms include domain-invariant feature learning~\cite{pmlr-v28-muandet13}, feature disentanglement~\cite{piratla2020efficient}, data augmentation~\cite{zhang2018mixup}, and meta-learning~\cite{li2018learning}. However, the majority of these algorithms are designed for unimodal data, such as images~\cite{li2017deeper} or time series data~\cite{oodrl}. With the increasing  need to process multimodal data in real-world applications~\cite{Damen2018EPICKITCHENS, nuscenes2019}, it has become imperative  to extend these methods to support  multimodal DG across a variety of modalities, including  audio-video~\cite{Kazakos_2019_ICCV,ZhangCVPR2022} and vision-language~\cite{clip,jia2021scaling}. In response to this challenge, several  approaches, such as 
RNA-Net~\cite{Planamente_2022_WACV} and SimMMDG~\cite{dong2023SimMMDG}, have been proposed to address the complexities of multimodal DG.

An inherent assumption in both DG and multimodal DG is the alignment of label spaces between source and target domains. However, real-world applications, such as autonomous driving~\cite{blum2019fishyscapes}, often feature target domains with novel categories not present in the source label space. As a result, the learned model may struggle with samples from these novel categories, significantly degrading the robustness of existing DG and multimodal DG methods. This setup, where the target domain contains unknown or open classes not seen in source domains,  is referred to as open-set DG. Several unimodal open-set DG approaches, including DAML~\cite{shu2021open} and MEDIC~\cite{wang2023generalizable}, have been developed  within the meta-learning framework to tackle this issue. 
CrossMatch~\cite{zhu2021crossmatch} utilizes an adversarial data augmentation strategy to generate auxiliary samples beyond the source label space. Yet, none specifically address the challenge of Multimodal Open-Set DG (MM-OSDG), which is the primary focus of this paper.
The goal of MM-OSDG is to train a model using data from several source domains across two or more modalities, enabling it to  effectively generalize to previously unseen target domains with the same modalities and including samples from unknown classes.  The key challenge of MM-OSDG is to  efficiently leverage complementary information from diverse modalities
to improve  generalization and open-class detection performance, areas where current unimodal open-set DG approaches fall short. We summarize the distinctions between our proposed MM-OSDG problem and various related problems in~\cref{tab:compare}.

\begin{table*}[t!]
\centering
\resizebox{0.8\linewidth}{!}{
\begin{tabular}{l|c|c|c}
\hline Task & Only one modality? & Need $\mathcal{C}_s=\mathcal{C}_t$? & Access to $\mathcal{D}_t$? \\
\hline Domain Adaptation~\cite{ganin2015unsupervised} & $\checkmark$ & $\checkmark$ & $\checkmark$ \\
Domain Generalization~\cite{carlucci2019domain} & $\checkmark$ & $\checkmark$ & $\times$ \\
Open-Set Domain Adaptation~\cite{liu2019separate} & $\checkmark$& $\times$ & $\checkmark$ \\
Open-Set Domain Generalization~\cite{shu2021open} & $\checkmark$ & $\times$ & $\times$ \\
Multimodal Domain Adaptation~\cite{Munro_2020_CVPR} & $\times$ & $\checkmark$ & $\checkmark$ \\
Multimodal Domain Generalization~\cite{dong2023SimMMDG} & $\times$ & $\checkmark$ & $\times$ \\
Multimodal Open-Set Domain Adaptation (proposed)&  $\times$& $\times$ & $\checkmark$ \\
Multimodal Open-Set Domain Generalization (proposed)& $\times$ & $\times$ & $\times$ \\
\hline
\end{tabular}
}
\vspace{0.1cm}
\caption{Illustration of the difference between the proposed Multimodal Open-Set DG and DA and other related tasks. $\mathcal{D}_t$ denote the target domains. $\mathcal{C}_s$ and $\mathcal{C}_t$ denote the label space of source and target domains, respectively.}
\label{tab:compare} 
\vspace{-1.0cm}
\end{table*}

To address the challenge of domain shift in unimodal setups, several studies have explored  the integration of self-supervision techniques for DG and DA~\cite{carlucci2019domain,bucci2020effectiveness}. This approach involves  addressing a self-supervised pretext task concurrently with  the primary supervised problem, which leads to the learning of resilient cross-domain features conducive to robust generalization. Additionally, recent research has demonstrated the utility of leveraging outputs from self-supervised models for anomaly detection,  facilitating the discrimination between normal and anomalous samples~\cite{bergman2020classification,golan2018deep} -- a scenario  similar to open-class detection. In particular, Rotation-based Open Set (ROS)~\cite{bucci2020effectiveness} explores rotation recognition as a self-supervised task for both unknown class detection and domain adaptation. Meanwhile, results from MOOD~\cite{li2023rethinking} illustrate that a reconstruction-based pretext task forces the network to learn the real data distribution of the in-distribution (ID) samples, thereby enlarging the divergence between the out-of-distribution (OOD) and ID samples.

Inspired by the success of the self-supervised pretext tasks in robust feature learning and OOD sample detection in unimodal setups, we propose a novel approach referred to as \textbf{\textit{MOOSA}} to address the \textbf{M}ultim\textbf{o}dal \textbf{O}pen-\textbf{S}et Domain Generalization and \textbf{A}daptation problems using multimodal self-supervised tasks. Specifically, we propose a generative task, termed Masked Cross-modal Translation, and a contrastive task, denoted as Multimodal Jigsaw Puzzles. These tasks are complementary to each other to effectively learn multimodal representative features conducive to both generalization and open-class detection. An entropy weighting mechanism is further introduced to balance the loss across different modalities. Additionally, our methodology is extended to handle the Multimodal Open-Set Domain Adaptation (MM-OSDA) problem in scenarios where unlabeled target data is available. The efficacy of the proposed approach is thoroughly validated through extensive experiments conducted on two multimodal DG benchmark datasets: EPIC-Kitchens~\cite{Damen2018EPICKITCHENS} and Human-Animal-Cartoon (HAC)~\cite{dong2023SimMMDG}. Our contributions can be summarized as follows:
\begin{itemize}
\setlength\itemsep{0.5em}
\item We delve into the unexplored area of Multimodal Open-Set Domain Generalization, a concept with significant implications for real-world applications. It requires a model trained on diverse source domains to generalize effectively to new, unseen target domains, which share the same modalities but include samples from previously unknown classes.
\item To tackle MM-OSDG, we introduce two complementary multimodal self-supervised tasks -- Masked Cross-modal Translation and Multimodal Jigsaw Puzzles -- along with an entropy weighting mechanism. We also extend our method to the novel Multimodal Open-Set Domain Adaptation setup.
\item The efficacy and versatility of our proposed approach are validated through extensive experiments conducted across MM-OSDG, MM-OSDA, and Multimodal Closed-Set DG settings.
\end{itemize}





\section{Related Work}

\noindent\textbf{Self-supervised Learning} generates supervisory signals based on the input data, alleviating the need for expensive manual labeling and improving performance across almost all types of downstream tasks, including domain generalization~\cite{carlucci2019domain} and anomaly detection~\cite{golan2018deep}. The majority of self-supervised tasks have so far focused on a single modality. 
Examples of such self-supervised tasks include predicting the relative position of image patches~\cite{doersch2015unsupervised}, colorizing  grayscale images~\cite{larsson2017colorization},  recognizing the rotation of an image~\cite{gidaris2018unsupervised}, or inpainting  removed patches~\cite{pathak2016context}. Recently, some approaches have been proposed to tackle the specific challenges of multimodal data, such as multimodal alignment~\cite{Munro_2020_CVPR}, relative norm alignment~\cite{Planamente_2022_WACV}, and cross-modal translation~\cite{dong2023SimMMDG}.

\noindent\textbf{Domain Generalization} entails training a model on multiple source domains to achieve robust generalization to previously unseen target domains. This is different from domain adaptation, where unlabeled target domain data is available during training.
Prior research~\cite{9782500} has identified three main categories of domain generalization methods:  data manipulation, representation learning, and learning strategies. 
Strategies involving data manipulation aim to enhance generalization performance by augmenting the diversity of training data~\cite{8202133,Zhou_Yang_Hospedales_Xiang_2020,zhang2018mixup}. 
Representation learning methods strive to acquire domain-invariant representations employing techniques such as domain-adversarial neural networks~\cite{ganin2016domain,li2018domain}, explicit feature distribution alignment~\cite{tzeng2014deep}, and instance normalization~\cite{pan2018two}. Additionally, various approaches leverage learning strategies to improve generalization performance, including meta-learning~\cite{li2018learning}, gradient operation~\cite{huang2020self}, and self-supervised learning~\cite{carlucci2019domain}.

\noindent\textbf{Open-Set DA and DG}
address scenarios where the target domain, with domain shifts, encompasses classes not present in the source domain~\cite{panareda2017open}. In the context of open-set DA, several approaches have been proposed to tackle this challenge, such as assigning target domain images to source categories while discarding unrelated target domain images~\cite{panareda2017open}, and employing adversarial training to separate unknown target samples~\cite{saito2018open}. Separate To Adapt (STA)~\cite{liu2019separate} introduces a progressive separation mechanism for unknown and known class samples. 
ROS~\cite{bucci2020effectiveness}  investigates the utility of rotation recognition for unknown class detection. 
Adjustment and Alignment (ANNA)~\cite{li2023adjustment} addresses semantic-level bias by incorporating front-door adjustment and decoupled causal alignment modules. For open-set DG, Domain-Augmented Meta-Learning (DAML)~\cite{shu2021open} adopts a meta-learning framework to learn open-domain generalizable representations and augment domains at both feature and label levels. CrossMatch~\cite{zhu2021crossmatch} employs an adversarial data augmentation strategy to generate auxiliary samples beyond the source label space. MEDIC~\cite{wang2023generalizable} considers gradient matching towards inter-domain and inter-class splits simultaneously to find a generalizable boundary balanced for all tasks. However, none of these approaches tackle the challenge posed by multiple modalities.

\noindent\textbf{Multimodal DA and DG}
tackle the challenges of DA and DG involving  multiple modalities. For instance, MM-SADA~\cite{Munro_2020_CVPR} proposes a self-supervised alignment approach coupled with adversarial alignment for multimodal DA, while Kim \etal~\cite{kim2021learning} leverage cross-modal contrastive learning to align cross-modal and cross-domain representations. Zhang \etal~\cite{ZhangCVPR2022} introduce an audio-adaptive encoder and an audio-infused recognizer to address domain shifts. RNA-Net~\cite{Planamente_2022_WACV} addresses the multimodal DG problem by introducing a relative norm alignment loss to balance audio and video feature norms. SimMMDG~\cite{dong2023SimMMDG} presents a universal framework for multimodal domain generalization, involving the separation of features within each modality into modality-specific and modality-shared components, along with constraints to facilitate meaningful representation learning. Different from prior efforts, our work represents the first attempt to address the Multimodal Open-Set DG and DA, scenarios that are both more practical and significantly more challenging.

\section{Methodology}
\begin{figure}[t!]
  \centering  \includegraphics[width=0.9\linewidth]{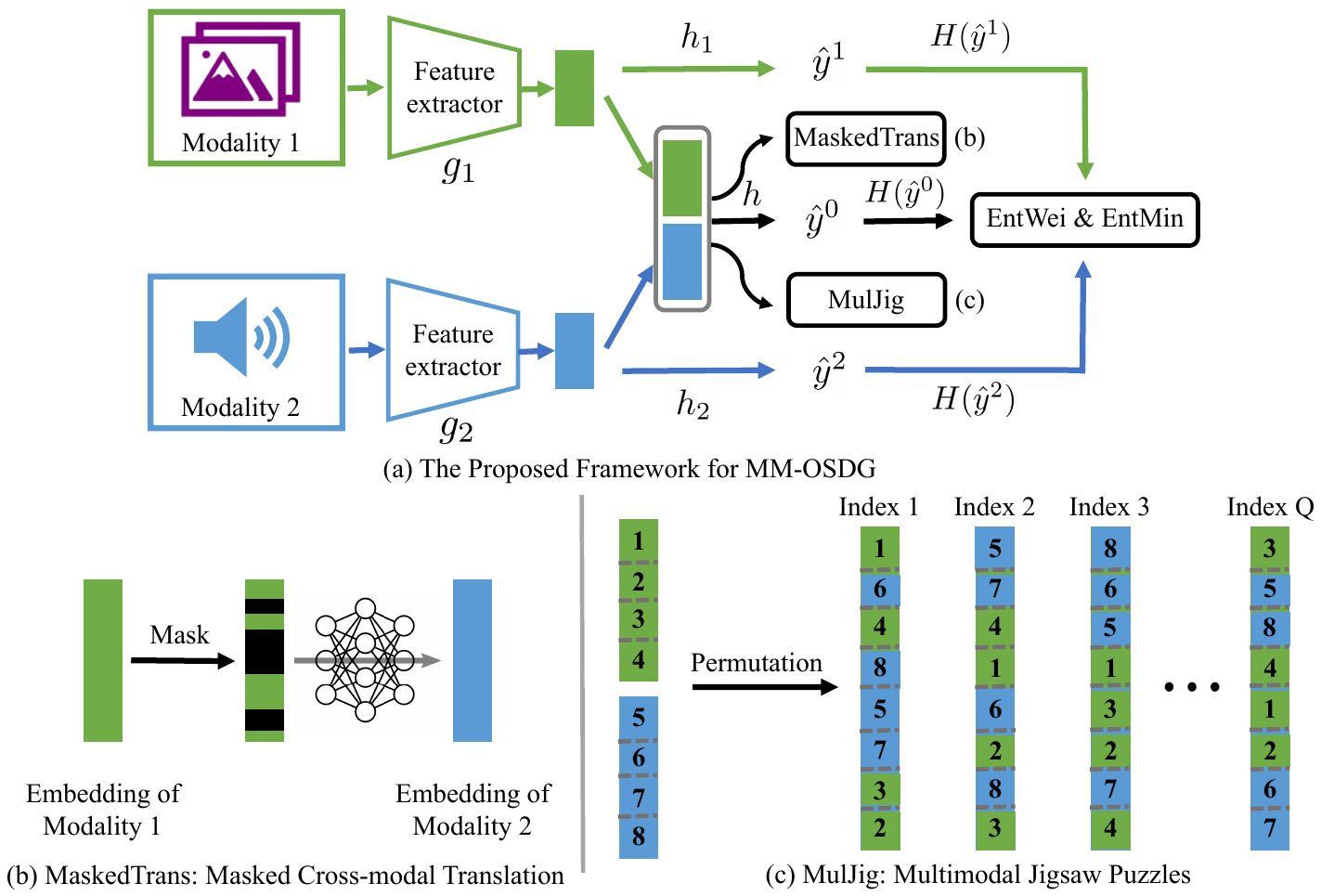}
\vspace{-0.3cm}
   \caption{Our proposed \textit{MOOSA} framework for MM-OSDG. EntWei \& EntMin: Entropy Weighting and Minimization.}
   \label{fig:framework}
\vspace{-0.4cm}
\end{figure}

In this section, we first introduce the formulation of the Multimodal Open-Set Domain Generalization problem. Subsequently, we present our novel framework designed to address this challenge. This framework incorporates the use of self-supervised pretext tasks, specifically the proposed Masked Cross-modal Translation and Multimodal Jigsaw Puzzles, complemented by an entropy weighting mechanism. Finally, we extend our framework to tackle  Multimodal Open-Set Domain Adaptation. A visual representation of our approach is depicted in~\cref{fig:framework}.

\subsection{Multimodal Open-Set Domain Generalization}
In the context of MM-OSDG, we work with several source domains, denoted as $\mathcal{S}=\lbrace \mathcal{D}_1,\mathcal{D}_2,...,\mathcal{D}_S\rbrace$, each sharing a common label space $\mathcal{C}$. We also consider an unseen target domain $\mathcal{T}=\lbrace \mathcal{D}_T\rbrace$ that introduces an extended label space $\mathcal{C} \cup \mathcal{U}$, with the stipulation  that  $\mathcal{C}$ and $\mathcal{U}$ do not overlap $\mathcal{C}\cap \mathcal{U}=\varnothing$. Each source domain, indexed by $s$ and comprising $N_s$ samples, is represented as $\mathcal{D}_s=\lbrace(\mathbf{x}^s_i,y^s_i)\rbrace_{i=1}^{N_s}$ following the distribution $ P^s_{XY}$. Here, $\mathbf{x}^s_i$ denotes the $i$-th sample from the sample space $\mathcal{X}$ and $y^s_i$ is the corresponding label, which belongs to the shared label space $\mathcal{C}$. 
Each sample instance $\mathbf{x}^s_i$ is comprised of $M$ modalities, denoted as $\mathbf{x}^s_i=\{(x^s_i)_k \mid k=1,\cdots,M\}$. 
The joint distributions between each pair of domains differ: $P^i_{XY} \ne P^j_{XY}, 1 \le i \ne j \le S$.
The objective of MM-OSDG is to develop a model  $f: \mathcal{X} \to \mathcal{Y}$ that leverages data from multiple  source domains  $S$ with a label space $\mathcal{C}$ and $M$ modalities. This model aims to seamlessly adapt  to unseen target domains $\mathcal{D}_T$ that include both known classes from $\mathcal{C}$ and unknown classes from $\mathcal{U}$, with the combined label space being
$\mathcal{C} \cup \mathcal{U}$. Consequently, the model is expected to accurately classify each target sample into the appropriate class from $\mathcal{C}$, or designate  it as “unknown” if it falls within $\mathcal{U}$. Importantly, during training,  data from the target domain $\mathcal{D}_T$ is  inaccessible, and there is a distribution shift between the source and the target domains,  $P^{T}_{XY} \ne P^s_{XY} \text{ for all }s \in \{1,\cdots,S\}$.

The predictive function $f$ comprises $M$ feature extractors $g_k(\cdot)$, a classifier $h(\cdot)$, and the Softmax function $\delta(\cdot)$. Each feature extractor $g_k(\cdot)$ is tasked with extracting a feature embedding $\mathbf{E}^{k}$ for its corresponding modality $k$, and the classifier $h(\cdot)$ takes the combined embeddings from all modalities as input to produce  the final output prediction $\hat{y}^0$, which can be written as:
\begin{equation}
\begin{split}
  \hat{y}^0 = \delta(f(\mathbf{x})) 
  &= \delta(h([g_1((x)_1), ..., g_k((x)_k), ..., g_M((x)_M)]))  \\
  &=  \delta(h([\mathbf{E}^{1}, ..., \mathbf{E}^{k}, ..., \mathbf{E}^{M}])) .
  \label{eqn:pred}
\end{split}
\end{equation}
We further include a separate classifier $h_k(\cdot)$ for each modality $k$, with the prediction from the $k$-th modality as $\hat{y}^k = \delta(h_k(g_k((x)_k)))$.

\subsection{Motivation: Self-supervised Pretext Tasks}
Self-supervised pretext tasks are commonly classified into generative and contrastive methods~\cite{liu2021self}. Contrastive approaches, such as context spatial relations~\cite{gidaris2018unsupervised,misra2020self} and instance discrimination~\cite{he2020momentum,chen2020simple}, focus on encouraging class-invariance between different instances~\cite{liu2021self}, thereby contributing to generalization. However, recent findings~\cite{saha2020hidden} indicate that contrastive methods may tend to learn specific patterns of categories, which are beneficial for classification, but less conducive to understanding intrinsic data distributions of in-distribution samples. In contrast, generative methods~\cite{devlin2018bert,he2022masked} force the network to learn the real data distribution of in-distribution samples during training, enhancing the divergence between in-distribution and out-of-distribution samples~\cite{li2023rethinking}, making them particularly valuable for open-class detection. Motivated by these observations, we propose using both generative and contrastive tasks to achieve  robust generalization and reliable open-class detection. Accordingly, we propose two novel multimodal self-supervised pretext tasks aligned with these two objectives. Multimodal self-supervised tasks are challenging, requiring the joint consideration of all modality properties to ensure meaningful outcomes.

\subsection{Generative Task: Masked Cross-modal Translation}
The cross-modal translation module introduced in SimMMDG~\cite{dong2023SimMMDG} can be regarded as an auto-encoding generative approach. It aims to ensure the meaningfulness of learned features by exploiting the implicit relationships inherent between the $M$ modalities within the same data instance. A multi-layer perceptron (MLP) is used to translate the feature embedding $\mathbf{E}$ across modalities. For instance, translating the embedding $\mathbf{E}^{i}$ of the $i$-th modality to the $j$-th modality using $MLP_{\mathbf{E}^{i} \rightarrow \mathbf{E}^{j}}$ results in the translated embedding $\mathbf{E}^{j}_{t}$. The aim is to minimize the $\ell_2$ distance between the real embedding of the $j$-th modality, $\mathbf{E}^{j}$, and the translated embedding, $\mathbf{E}^{j}_{t}$, defining the cross-modal translation loss as:
\begin{equation}
  \mathcal{L}_{trans}
  =\frac{1}{M(M-1)} \sum_{i=1}^{M} \sum_{j \neq i} ||MLP_{\mathbf{E}^{i} \rightarrow \mathbf{E}^{j}}(\mathbf{E}^{i}) - \mathbf{E}^{j}||_2^2 .
  \label{eqn:trans_loss}
\end{equation}

Although cross-modal translation has demonstrated effectiveness in various multimodal DG tasks~\cite{dong2023SimMMDG}, inspired by the masked image modeling pretext task~\cite{bao2021beit} and its application in OOD detection~\cite{li2023rethinking}, we propose masked cross-modal translation tailored to better align with the MM-OSDG task. Instead of directly translating $\mathbf{E}^{i}$ to $\mathbf{E}^{j}$, we randomly mask a percentage of $\mathbf{E}^{i}$ and subsequently translate this masked embedding, denoted as $Mask(\mathbf{E}^{i})$, to the target embedding, as depicted in~\cref{fig:framework} (b). The masking operation, $Mask(\mathbf{E}^{i})$, randomly sets a $\beta$ percentage of elements in $\mathbf{E}^{i}$ to $0$. The masked cross-modal translation loss is then defined based on this operation as follows:
\begin{equation}
  \mathcal{L}_{MaskedTrans}
  =\frac{1}{M(M-1)} \sum_{i=1}^{M} \sum_{j \neq i} ||MLP_{\mathbf{E}^{i} \rightarrow \mathbf{E}^{j}}(Mask(\mathbf{E}^{i}) )- \mathbf{E}^{j}||_2^2 .
  \label{eqn:mask_trans_loss}
\end{equation}
With this masking procedure, the model learns to infer the complete embeddings of the target modality from the remaining embeddings of the source modality. This procedure enables  the model to leverage  intrinsic data distributions across diverse modalities as priors, rather than solely focusing on discerning different patterns among categories during  the classification process. By benefiting from these intrinsic priors, the model achieves  a more representative feature representation of the known classes, thereby increasing  the divergence between the known and unknown samples. This observation is further supported by the visualization results in~\cref{fig:score}.


\subsection{Contrastive Task: Multimodal Jigsaw Puzzles}
Context-based spatial relation prediction~\cite{gidaris2018unsupervised,doersch2015unsupervised} falls within the category of contrastive objectives~\cite{liu2021self}.
Previous studies ~\cite{carlucci2019domain,noroozi2016unsupervised} have explored learning visual representations through solving Jigsaw puzzles, a task that involves recovering an original image from its shuffled parts. 
In this work, we extend the Jigsaw puzzle-solving task to multimodal scenarios. 
Given the heterogeneity of different modalities, shuffling them in the input space and then randomly combining them poses a challenge. Therefore, our approach involves shuffling in the embedding space for all modalities. Given the embedding $\mathbf{E}^{i}$ of the $i$-th modality, we split it into $P$ parts as $\mathbf{E}^{i} = [\mathbf{E}^{i,1}, \mathbf{E}^{i,2}, ...,  \mathbf{E}^{i,P}]$ with equal length. Similarly, for the $j$-th modality, we have $\mathbf{E}^{j} = [\mathbf{E}^{j,1}, \mathbf{E}^{j,2}, ...,  \mathbf{E}^{j,P}]$. 
Combining all $M \cdot P$ split parts from $M$ modalities, we randomly shuffle them to generate different permutations.
For example, with two modalities $i$ and $j$ and by setting $P$ to $2$, one possible permutation is $\mathbf{\widetilde{E}}^{p} = [\mathbf{E}^{j,1}, \mathbf{E}^{i,2}, \mathbf{E}^{j,2},  \mathbf{E}^{i,1}]$. The shuffling is performed across modalities, which makes the task particularly challenging.
Out of the $(M \cdot P)!$ possible permutations, we randomly select a set of $Q$ elements and assign an index to each entry as the label. Then, we can define an auxiliary classification task on each sample instance as $\{(\mathbf{\widetilde{E}}^{p},p)\}_{p=1}^{Q}$, where $\mathbf{\widetilde{E}}^{p}$ indicates the recomposed embeddings and $p \in \{1,\ldots,Q\}$ denotes the corresponding permutation index. To achieve this, we aim to minimize the multimodal jigsaw loss  $\mathcal{L}_{MulJig}(h_p(\mathbf{\widetilde{E}}^{p}),p)$, were $h_p$ is the classifier for permutation recognition, and $\mathcal{L}_{MulJig}$ is the standard cross-entropy loss. \cref{fig:framework} (c) provides  an illustration of this process.

By solving the multimodal Jigsaw puzzles problem, the model must not only to recover the correct order of the shuffled embeddings for each modality but also discern the source modality of each shuffled part. This process enables the model to learn multimodal representative features conducive to robust generalization. 

\subsection{Entropy Weighting and Minimization}
Distinct modalities may contribute differently to the final predictions under various circumstances. For instance, in recognizing the "running" action within a noisy gym, video data may be more reliable than audio data for the final prediction. Conversely, in scenarios characterized by low environmental light or partial shading of the camera, audio input may prove more dependable. To address the need for balancing the contributions of different modalities during training, we propose employing the prediction entropy of each modality to weigh their respective losses. The entropy $H(\hat{y})$ of model predictions $\hat{y}$ is calculated as $H(\hat{y}) = -\sum_c p(\hat{y}_c) \log p(\hat{y}_c)$, where $\hat{y}_c$ represents the probability of class $c$. Entropy serves as a metric for prediction confidence, with higher confidence yielding lower entropy values. For each prediction $\hat{y}^k$ from the $k$-th modality, we calculate an entropy  $H(\hat{y}^k)$ and a cross-entropy loss $\mathcal{L}_{cls}^k = CE(\hat{y}^k, {y})$. The objective is to assign less weight to the modality with higher entropy in contributing to the final prediction loss. Given $M+1$ entropy values $H(\hat{y}^0)$, $H(\hat{y}^1)$, ..., $H(\hat{y}^M)$, the weight of the $k$-th prediction is calculated by:
\begin{equation}
w_k=\frac{e^{-H(\hat{y}^k)/T}}{\sum_{i=0}^M e^{-H(\hat{y}^i)/T}} ,
\end{equation}
where $T$ is the temperature. Then, the final prediction loss is defined as:
\begin{equation}
  \mathcal{L}_{cls}
  =\sum_{k} w_k \mathcal{L}_{cls}^k.
  \label{eqn:l_all}
\end{equation}
Similar to previous work~\cite{wang2020tent}, we also add an entropy minimization loss as: 
\begin{equation}
\mathcal{L}_{EntMin} = \sum_{k} H(\hat{y}^k).
\end{equation}

\subsection{Final Loss and Inference}
The final loss is obtained as the weighted sum of the previously defined losses:
\begin{equation}
  \mathcal{L}
  =\mathcal{L}_{cls} + \alpha_{1} \mathcal{L}_{MaskedTrans}+ \alpha_{2} \mathcal{L}_{MulJig}+ \alpha_{3} \mathcal{L}_{EntMin} ,
  \label{eqn:l_all2}
\end{equation}
where $\mathcal{L}_{cls}$ is the cross-entropy loss for classification. The hyperparameters $\alpha_{\text{1}}$, $\alpha_{\text{2}}$, and $\alpha_{\text{3}}$ control the relative importance of the Masked Cross-modal Translation, Multimodal Jigsaw Puzzles, and Entropy Minimization, respectively.

During inference, we use $\hat{y}^0$ as the final prediction. Following the approach outlined in previous research~\cite{you2019universal}, we employ a threshold on the prediction confidence. Samples with a confidence level lower than the specified threshold are subsequently labeled as belonging to an open class.


\subsection{Extension to Multimodal Open-Set Domain Adaptation}
Different from MM-OSDG, in MM-OSDA, we have access to sample instances $\mathcal{D}_T=\lbrace(\mathbf{x}^T_i)\rbrace_{i=1}^{N_T}$ from the target domain during training. However, the labels $y^T_i$ for each $\mathbf{x}^T_i$ are not available, where $y^T_i \in \mathcal{C} \cup \mathcal{U}$, encompassing both known and unknown classes. The first step is to distinguish the known samples ($y^T_i \in \mathcal{C}$) and unknown samples ($y^T_i \in \mathcal{U}$) from $\mathcal{D}_T$ to mitigate the risk of negative transfer. Given a sample $\mathbf{x}^T$ from the target domain, we calculate a prediction $\hat{y}^0 = \delta(f(\mathbf{x^T}))$ and use $max(\hat{y}^0)$ as the prediction confidence to identify unknown classes. The assumption is that the model exhibits less confidence in predicting unknown samples compared to known ones. A target sample is classified as unknown if $max(\hat{y}^0)$ falls below a threshold $\tau$, where we typically set $\tau = 0.5$. 

After filtering out all unknown samples from the target domain, we apply our self-supervised pretext tasks to the remaining known samples $\widetilde{\mathcal{D}}_T$ to align the source and target domains. For a given target sample $\widetilde{\mathbf{x}}^T$ from $\widetilde{\mathcal{D}}_T$, we first calculate the embedding for each modality as $[\widetilde{\mathbf{E}}^{1}, ..., \widetilde{\mathbf{E}}^{k}, ..., \widetilde{\mathbf{E}}^{M}]$ and then use $\mathcal{L}_{MaskedTrans}$ and $\mathcal{L}_{M ulJig}$ to calculate the losses associated with different self-supervised pretext tasks. Additionally, we incorporate $\mathcal{L}_{EntMin}$ into the prediction of target samples. Finally, we can redefine our loss under the MM-OSDA setting as:
\begin{equation}
\begin{split}
  \mathcal{L}
  =\mathcal{L}_{cls} + \alpha_{1} (\mathcal{L}_{MaskedTrans}^s+\mathcal{L}_{MaskedTrans}^t)  + \alpha_{2} (\mathcal{L}_{MulJig}^s+\mathcal{L}_{MulJig}^t) \\  + \alpha_{3} (\mathcal{L}_{EntMin}^s+\mathcal{L}_{EntMin}^t) ,
  \label{eqn:l_all3}
\end{split}
\end{equation}
where $\mathcal{L}^s$ and $\mathcal{L}^t$ correspond to different losses applied to source and target domain data respectively.

\section{Experiments}
\subsection{Experimental Setting}
\noindent\textbf{{Dataset.}} We use the EPIC-Kitchens~\cite{Damen2018EPICKITCHENS} and HAC~\cite{dong2023SimMMDG} datasets for experiments and adjust these to the open-set setups. We follow the experimental protocol used in previous research~\cite{dong2023SimMMDG}. The EPIC-Kitchens dataset includes eight actions (‘put’, ‘take’, ‘open’, ‘close’, ‘wash’, ‘cut’, ‘mix’, and ‘pour’) recorded across three distinct kitchens, constituting three separate domains D1, D2, and D3. The HAC dataset comprises seven actions (‘sleeping’, ‘watching TV’, ‘eating’, ‘drinking’, ‘swimming’, ‘running’, and ‘opening door’) performed by humans, animals, and cartoon figures, forming three distinct domains H, A, and C. Both datasets feature three modalities, including video, audio, and optical flow. For both datasets, we organized the action names in alphabetic order, designating the first class as unknown and the remaining classes as known. Additionally, we explored the impact of different levels of openness in the ablation study.

\noindent\textbf{{Evaluation Metrics.}}
For performance evaluation, we employ three widely recognized evaluation metrics commonly used in OSDA literature \cite{bucci2020effectiveness,liu2019separate,li2023adjustment}. In OSDA, rather than using average per-class accuracy $OS$, we present results using the harmonic mean $HOS$. The $HOS$ is calculated as $\frac{2\times OS^*\times UNK}{OS^* + UNK}$, where $OS^*$ represents the accuracy in known categories, and $UNK$ represents the accuracy in unknown classes. The $HOS$ metric offers a balance between accuracy in known and unknown categories, providing a more nuanced evaluation than OS alone. This distinction is particularly important in scenarios where the accuracy of unknown classes is notably low, emphasizing the significance of effective detection of unknown categories. The $HOS$ is alternatively denoted as H-score in some paper~\cite{shu2021open,wang2023generalizable}.

\noindent\textbf{{Baselines.}}
As the first to address the Multimodal Open-Set DG and DA problems, we benchmark our method against several state-of-the-art (SOTA) Multimodal DG and DA approaches, including MM-SADA~\cite{Munro_2020_CVPR}, RNA-Net~\cite{Planamente_2022_WACV}, and SimMMDG~\cite{dong2023SimMMDG}. Additionally, MEDIC~\cite{wang2023generalizable}, the SOTA unimodal Open-Set DG method based on the meta-learning framework, is extended to multimodal scenarios for a comprehensive comparison. The DeepAll approach involves  feeding all data from source domains to the network without employing any domain generalization strategies.
For all baselines, experiments are conducted using the official code provided by the respective authors, and we apply the same prediction confidence-based method as in this paper to detect unknown classes. More implementation details are provided in~\cref{imple}.

\subsection{Results}

\noindent\textbf{{Multimodal Open-set DG.}}
\begin{table}[t!]
\centering
\resizebox{0.85\textwidth}{!}{
\begin{tabular}{l@{~~~~~}cc ccc ccc ccc ccc ccc}
\hline

 & &\multicolumn{3}{c|}{\textbf{Modality} } &\multicolumn{3}{c|}{D2, D3 $\rightarrow$ D1 } & \multicolumn{3}{c|}{D1, D3 $\rightarrow$ D2 } & \multicolumn{3}{c|}{D1, D2 $\rightarrow$ D3 }  &  \multicolumn{3}{c}{\textit{Mean} }
 \\
        &  & Video  & Audio & \multicolumn{1}{c|}{Flow} & OS* & UNK & \multicolumn{1}{c|}{\textbf{\underline{HOS}}} &  OS* & UNK &  \multicolumn{1}{c|}{\textbf{\underline{HOS}}}  & OS* & UNK &  \multicolumn{1}{c|}{\textbf{\underline{HOS}}}  & OS* & UNK &  \multicolumn{1}{c}{\textbf{\underline{HOS}}} \\
\hline
{DeepAll} &  & $\checkmark$& $\checkmark$&  \multicolumn{1}{c|}{} & 38.90 & 50.00 & \multicolumn{1}{c|}{43.76}  & 47.69 & 44.64 & \multicolumn{1}{c|}{46.12}  & 52.16 & 42.25 & \multicolumn{1}{c|}{46.69}  & 46.25 &45.63  & \multicolumn{1}{c}{45.52}  \\

{MM-SADA~\cite{Munro_2020_CVPR}} &  & $\checkmark$& $\checkmark$&  \multicolumn{1}{c|}{} & 35.41 &  50.00 & \multicolumn{1}{c|}{41.46}  & 47.12 & 50.00 & \multicolumn{1}{c|}{48.52}  &51.16  & 46.48 & \multicolumn{1}{c|}{48.71}  & 44.56 & 48.83 & \multicolumn{1}{c}{46.23}  \\

{RNA-Net~\cite{Planamente_2022_WACV}} &  & $\checkmark$& $\checkmark$&  \multicolumn{1}{c|}{} &39.90  & 38.24 & \multicolumn{1}{c|}{39.05}  & 43.80 & 53.57 & \multicolumn{1}{c|}{48.20}  & 46.29 & 59.15 & \multicolumn{1}{c|}{51.94}  & 43.33 & 50.32 & \multicolumn{1}{c}{46.40}  \\

{MEDIC~\cite{wang2023generalizable}} &  & $\checkmark$& $\checkmark$&  \multicolumn{1}{c|}{} & 38.40 & 55.88 & \multicolumn{1}{c|}{45.52}  & 34.44&  69.64& \multicolumn{1}{c|}{46.09}   &39.98  &59.15  & \multicolumn{1}{c|}{47.71}   &37.61  & 61.56 & \multicolumn{1}{c}{46.44}  \\

{SimMMDG~\cite{dong2023SimMMDG}} &  & $\checkmark$& $\checkmark$&  \multicolumn{1}{c|}{} &  34.41& 70.59 & \multicolumn{1}{c|}{46.27}  & 46.83 & 51.79 & \multicolumn{1}{c|}{49.18}  & 43.19 & 63.38 & \multicolumn{1}{c|}{51.37}  & 41.48 & 61.92 & \multicolumn{1}{c}{48.94}  \\

{MOOSA (ours)} &  & $\checkmark$& $\checkmark$&  \multicolumn{1}{c|}{} & 41.90 &82.35& \multicolumn{1}{c|}{\textbf{55.54}}  & 45.53 & 64.29 & \multicolumn{1}{c|}{\textbf{53.31}}  & 45.52 & 69.01& \multicolumn{1}{c|}{\textbf{54.85}}  &44.32  & 71.88 & \multicolumn{1}{c}{\textbf{54.57}} \\

\hline

{DeepAll} &  & $\checkmark$& &  \multicolumn{1}{c|}{$\checkmark$} &47.88  & 47.06 & \multicolumn{1}{c|}{47.47}  & 48.42 & 44.64 & \multicolumn{1}{c|}{46.45}  & 50.28 & 36.62 & \multicolumn{1}{c|}{42.38}  & 48.86 & 42.77 & \multicolumn{1}{c}{45.43}  \\

{MM-SADA~\cite{Munro_2020_CVPR}} &  & $\checkmark$& &  \multicolumn{1}{c|}{$\checkmark$} & 44.39 & 47.06 & \multicolumn{1}{c|}{45.69}  & 40.20 &48.21  & \multicolumn{1}{c|}{43.84}  & 45.96 & 47.89 & \multicolumn{1}{c|}{46.90}  & 43.52 & 47.72 & \multicolumn{1}{c}{45.48}  \\

{RNA-Net~\cite{Planamente_2022_WACV}} &  & $\checkmark$& &  \multicolumn{1}{c|}{$\checkmark$} & 48.13 & 50.00 & \multicolumn{1}{c|}{49.05}  & 50.72 & 48.21 & \multicolumn{1}{c|}{49.44}  & 51.61 & 39.44 & \multicolumn{1}{c|}{44.71}  & 50.15 & 45.88 & \multicolumn{1}{c}{47.73}  \\

{MEDIC~\cite{wang2023generalizable}} &  & $\checkmark$& &  \multicolumn{1}{c|}{$\checkmark$} & 44.14 & 61.76 & \multicolumn{1}{c|}{51.49}  & 43.37& 60.71 & \multicolumn{1}{c|}{50.60}   & 40.42 & 56.34 & \multicolumn{1}{c|}{47.07}   & 42.64 & 59.60 & \multicolumn{1}{c}{49.72}  \\

{SimMMDG~\cite{dong2023SimMMDG}} &  & $\checkmark$& &  \multicolumn{1}{c|}{$\checkmark$} &42.14  & 67.65 & \multicolumn{1}{c|}{51.93}  & 41.93 & 78.57 & \multicolumn{1}{c|}{54.68}  & 41.75 & 64.79 & \multicolumn{1}{c|}{50.78}  & 41.94 & 70.34 & \multicolumn{1}{c}{52.46}  \\

{MOOSA (ours)} &  & $\checkmark$& &  \multicolumn{1}{c|}{$\checkmark$} & 45.14 &61.76  & \multicolumn{1}{c|}{\textbf{52.16}}  & 50.00 & 62.50 & \multicolumn{1}{c|}{\textbf{55.56}}  &  47.84& 54.93 & \multicolumn{1}{c|}{\textbf{51.14}}  & 47.66 & 59.73 & \multicolumn{1}{c}{\textbf{52.95}}  \\

\hline

{DeepAll} &  & &$\checkmark$ &  \multicolumn{1}{c|}{$\checkmark$} & 40.90 & 44.12 & \multicolumn{1}{c|}{42.45}  & 46.25 & 37.50 & \multicolumn{1}{c|}{41.42}  & 38.32 & 46.48 & \multicolumn{1}{c|}{42.01}  & 41.82 & 42.70 & \multicolumn{1}{c}{41.96}  \\

{MM-SADA~\cite{Munro_2020_CVPR}} &  & &$\checkmark$ &  \multicolumn{1}{c|}{$\checkmark$} & 38.90 & 47.06 & \multicolumn{1}{c|}{42.59}  & 45.97 & 42.86 & \multicolumn{1}{c|}{44.36}  & 45.52 & 40.85 & \multicolumn{1}{c|}{43.05}  & 43.46 & 43.59 & \multicolumn{1}{c}{43.33}  \\

{RNA-Net~\cite{Planamente_2022_WACV}} &  &&  $\checkmark$&  \multicolumn{1}{c|}{$\checkmark$} & 37.16 & 52.94 & \multicolumn{1}{c|}{43.67}  & 46.69 & 46.43 & \multicolumn{1}{c|}{46.56}  &47.51  & 40.85 & \multicolumn{1}{c|}{43.93}  & 43.79 &46.74  & \multicolumn{1}{c}{44.72}  \\

{MEDIC~\cite{wang2023generalizable}} &  & & $\checkmark$ &  \multicolumn{1}{c|}{$\checkmark$} &  44.39& 47.06 & \multicolumn{1}{c|}{45.69}  & 38.47& 60.71 & \multicolumn{1}{c|}{47.10}   &40.53  & 49.30 & \multicolumn{1}{c|}{44.49}   & 41.13 & 52.36 & \multicolumn{1}{c}{45.76}  \\

{SimMMDG~\cite{dong2023SimMMDG}} &  & & $\checkmark$&  \multicolumn{1}{c|}{$\checkmark$} & 43.64& 52.94 & \multicolumn{1}{c|}{ 47.84}  &  46.54& 67.86 & \multicolumn{1}{c|}{55.21}  &36.88  & 60.56 & \multicolumn{1}{c|}{45.84}  & 42.35 & 60.45 & \multicolumn{1}{c}{49.63}  \\

{MOOSA (ours) } &  & &$\checkmark$ &  \multicolumn{1}{c|}{$\checkmark$} & 42.64 & 55.88 & \multicolumn{1}{c|}{\textbf{48.37}}  &  46.69& 75.00 & \multicolumn{1}{c|}{\textbf{57.55}}  &  42.19& 57.75 & \multicolumn{1}{c|}{\textbf{48.76}}  & 43.84 & 62.88 & \multicolumn{1}{c}{\textbf{51.56}}  \\

\hline

{DeepAll} &  & $\checkmark$&$\checkmark$ &  \multicolumn{1}{c|}{$\checkmark$} & 39.15 & 50.00 & \multicolumn{1}{c|}{43.92}  & 52.45 &  35.71& \multicolumn{1}{c|}{42.49}  & 53.82 &  39.44& \multicolumn{1}{c|}{45.52}  &48.47  & 41.72 & \multicolumn{1}{c}{43.98}  \\

{MM-SADA~\cite{Munro_2020_CVPR}} &  & $\checkmark$&$\checkmark$ &  \multicolumn{1}{c|}{$\checkmark$} & 38.90 & 41.18 & \multicolumn{1}{c|}{40.01}  & 45.68 & 51.79 & \multicolumn{1}{c|}{48.54}  & 44.41 &54.93  & \multicolumn{1}{c|}{49.11}  &43.00  & 49.30 & \multicolumn{1}{c}{45.89}  \\

{RNA-Net~\cite{Planamente_2022_WACV}} &  & $\checkmark$&$\checkmark$ &  \multicolumn{1}{c|}{$\checkmark$} & 40.15 & 50.00 & \multicolumn{1}{c|}{44.54}  &  48.85&  44.64& \multicolumn{1}{c|}{46.65}  & 48.39 & 49.30 & \multicolumn{1}{c|}{48.84}  & 45.80 & 47.98 & \multicolumn{1}{c}{46.68}  \\

{MEDIC~\cite{wang2023generalizable}} &  &$\checkmark$& $\checkmark$ &  \multicolumn{1}{c|}{$\checkmark$} & 43.64 & 64.71 & \multicolumn{1}{c|}{52.13}  & 47.84& 64.29 & \multicolumn{1}{c|}{54.86}   &  45.29& 59.15 & \multicolumn{1}{c|}{51.30}   & 45.59 & 62.72 & \multicolumn{1}{c}{52.76}  \\

{SimMMDG~\cite{dong2023SimMMDG}} &  & $\checkmark$& $\checkmark$&  \multicolumn{1}{c|}{$\checkmark$} & 40.15 & 73.53 & \multicolumn{1}{c|}{51.94}  & 45.68 & 78.57 & \multicolumn{1}{c|}{57.77}  & 39.09 & 78.87 & \multicolumn{1}{c|}{52.27}  & 41.64 & 76.99 & \multicolumn{1}{c}{53.99}  \\

{MOOSA (ours)} &  & $\checkmark$&$\checkmark$ &  \multicolumn{1}{c|}{$\checkmark$} & 46.13 &73.52  & \multicolumn{1}{c|}{\textbf{56.70}}  &53.03  &  67.86& \multicolumn{1}{c|}{\textbf{59.53}}  & 48.84 & 60.56 & \multicolumn{1}{c|}{\textbf{54.07}}  & 49.33 & 67.31 & \multicolumn{1}{c}{\textbf{56.77}}  \\

\hline
\end{tabular} 
}
\vspace{0.1cm}
\caption{Multimodal \textbf{Open-set DG} with different combinations of modalities on EPIC-Kitchens dataset.}
\vspace{-0.7cm}
\label{tab:epic-open} 
\end{table}
\begin{table}[t!]
\centering
\resizebox{0.85\textwidth}{!}{
\begin{tabular}{l@{~~~~~}cc ccc ccc ccc ccc ccc}
\hline

 & &\multicolumn{3}{c|}{\textbf{Modality} } &\multicolumn{3}{c|}{A, C $\rightarrow$ H } & \multicolumn{3}{c|}{H, C $\rightarrow$ A } & \multicolumn{3}{c|}{H, A $\rightarrow$ C }  &  \multicolumn{3}{c}{\textit{Mean} }
 \\
        &  & Video  & Audio & \multicolumn{1}{c|}{Flow} & OS* & UNK & \multicolumn{1}{c|}{\textbf{\underline{HOS}}} &  OS* & UNK &  \multicolumn{1}{c|}{\textbf{\underline{HOS}}}  & OS* & UNK &  \multicolumn{1}{c|}{\textbf{\underline{HOS}}}  & OS* & UNK &  \multicolumn{1}{c}{\textbf{\underline{HOS}}} \\
\hline

{DeepAll} &  & $\checkmark$&$\checkmark$ &  \multicolumn{1}{c|}{$\checkmark$} & 66.02 &71.72  & \multicolumn{1}{c|}{68.75}  & 74.51 & 54.14 & \multicolumn{1}{c|}{62.71}  & 44.59 & 55.56 & \multicolumn{1}{c|}{49.48}  &  61.71& 60.47 & \multicolumn{1}{c}{60.31}  \\

{MM-SADA~\cite{Munro_2020_CVPR}} &  & $\checkmark$&$\checkmark$ &  \multicolumn{1}{c|}{$\checkmark$} & 60.13 & 72.22 & \multicolumn{1}{c|}{65.63}  & 76.46 & 52.63 & \multicolumn{1}{c|}{62.35}  &  40.64  & 56.35 & \multicolumn{1}{c|}{47.23}  &59.08  & 60.40 & \multicolumn{1}{c}{58.40}  \\

{RNA-Net~\cite{Planamente_2022_WACV}} &  & $\checkmark$&$\checkmark$ &  \multicolumn{1}{c|}{$\checkmark$} & 70.31 & 67.68 & \multicolumn{1}{c|}{68.97}  & 72.06 & 51.88 & \multicolumn{1}{c|}{60.33}  &46.99  &55.56 & \multicolumn{1}{c|}{50.91}  & 63.12 & 58.37 & \multicolumn{1}{c}{60.07}  \\

{MEDIC~\cite{wang2023generalizable}} &  &$\checkmark$& $\checkmark$ &  \multicolumn{1}{c|}{$\checkmark$} & 56.18 & 77.78 & \multicolumn{1}{c|}{65.24}  &58.47 & 69.17 & \multicolumn{1}{c|}{63.37}   & 42.93 & 61.11 & \multicolumn{1}{c|}{50.43}   &  52.53& 69.35 & \multicolumn{1}{c}{59.68}  \\

{SimMMDG~\cite{dong2023SimMMDG}} &  & $\checkmark$& $\checkmark$&  \multicolumn{1}{c|}{$\checkmark$} & 62.83 & 77.27 & \multicolumn{1}{c|}{69.30}  & 63.78 & 75.19 & \multicolumn{1}{c|}{69.01}  & 42.41 & 64.29 & \multicolumn{1}{c|}{51.11}  &56.34  & 72.25 & \multicolumn{1}{c}{63.14}  \\

{MOOSA (ours)} &  & $\checkmark$&$\checkmark$ &  \multicolumn{1}{c|}{$\checkmark$} & 66.69 & 78.79 & \multicolumn{1}{c|}{\textbf{72.24}}  &66.36 & 75.94 & \multicolumn{1}{c|}{\textbf{70.83}}  &  45.01& 62.70 & \multicolumn{1}{c|}{\textbf{52.40}}  & 59.35& 72.48 & \multicolumn{1}{c}{\textbf{65.16}}\\

\hline
\end{tabular} 
}
\vspace{0.1cm}
\caption{Multimodal \textbf{Open-set DG} with all modalities on HAC dataset. The results for different combinations of any two modalities are in the Appendix.}
\vspace{-0.9cm}
\label{tab:hac-open-less} 
\end{table}
\cref{tab:epic-open} and \cref{tab:hac-open-less} present the results under the multimodal open-set DG setting, where the model is trained on several source domains with multiple modalities and tested on a single target domain with open classes. To verify the adaptability of our framework to various modalities, we conduct experiments by combining any two modalities, as well as all three modalities. The results on the EPIC-Kitchens dataset demonstrate that \textit{MOOSA} consistently outperforms all baselines by a substantial margin across all cases, yielding average improvements in $HOS$ of up to $5.63\%$. Notably, our performance further improves when combining all three modalities, surpassing the results achieved with any two modalities. Conversely, some baseline methods fail to exhibit improved results with an increased number of modalities, suggesting an inability to effectively leverage the complementary information between modalities.
To validate the versatility of \textit{MOOSA}, we extend the experiments to the HAC dataset, with results consistent with those obtained on the EPIC-Kitchens dataset, as detailed in~\cref{tab:hac-open-less} and in~\cref{tab:hac-open}. Our method consistently outperforms all baselines by a significant margin in most cases, yielding average improvements in terms of $HOS$ of up to $3.37\%$.

\begin{wraptable}{r}{0.5\textwidth}
\centering
\resizebox{\linewidth}{!}{
\begin{threeparttable}
\begin{tabular}{lcccccccc}
\toprule
\qquad\qquad\;\; Source:&  \multicolumn{2}{c}{D1}& \multicolumn{2}{c}{D2}& \multicolumn{2}{c}{D3}\\
\cmidrule(lr){2-3} \cmidrule(lr){4-5} \cmidrule(lr){6-7} 
\textbf{Method} \; Target: &  D2 &  D3 &  D1 &  D3& D1&  D2  & \textit{Mean}\\

\midrule
DeepAll &  40.63 & 39.92 & 37.34 & 38.17 & 33.35 & 41.00 & 38.40 \\
MM-SADA~\cite{Munro_2020_CVPR} &  40.78 & 38.05 & 33.23 & 36.14 & 35.48 & 41.55 & 37.54 \\
RNA-Net~\cite{Planamente_2022_WACV} &40.95   & 40.86 & 32.61 & 38.39 & 32.01 & 44.62 & 38.24 \\
SimMMDG~\cite{dong2023SimMMDG}&  43.28 & 43.67 & 43.91 & 48.68 & 45.48 & \textbf{51.18} & 46.03 \\
MOOSA (ours)&  \textbf{49.04} & \textbf{46.55} & \textbf{44.01} & \textbf{51.89} &\textbf{50.39}  & 49.00 & \textbf{48.48} \\

\bottomrule
\end{tabular}

\end{threeparttable}
}
\caption{Multimodal \textbf{Single-source Open-set DG} with video and audio modalities on EPIC-Kitchens dataset. The $HOS$ value is reported.}

\vspace{-0.7cm}
\label{tab:epic-ssdg-appen}
\end{wraptable} 

\noindent\textbf{{Multimodal Single-source Open-set DG.}}
The inclusion of domain labels is not required within our framework. This characteristic makes our method immediately applicable to multimodal single-source open-set DG without necessitating modifications, where the model is trained exclusively on a single source domain and subsequently tested across multiple target domains. The $HOS$ results (\cref{tab:epic-ssdg-appen}) within the Multimodal Single-source Open-set DG paradigm demonstrate robust generalization to unseen domains of \textit{MOOSA}, with an average $HOS$ improvement of up to $2.45 \%$.

\noindent\textbf{{Multimodal Open-set DA.}} In this setting, unlabeled data from the target domain is available during training. As we are the first to address the MM-OSDA problem, we extend several methods to this setup and consider them as baselines. For MM-SADA~\cite{Munro_2020_CVPR}, we employ the within-modal adversarial alignment loss to align the distribution of each modality between source and target domains. In the case of RNA-Net~\cite{Planamente_2022_WACV}, we apply the relative norm alignment loss on the target domain data. Additionally, for SimMMDG~\cite{dong2023SimMMDG}, we incorporate their contrastive loss, distance loss, and cross-modal translation loss to the target data to align the distribution across domains. As depicted in~\cref{tab:epic-mmda}, our method achieves competitive performance compared to the best baseline,  
yielding an improvement of $4.15\%$ on $HOS$.

\begin{table}[t!]
\centering
\resizebox{\textwidth}{!}{
\begin{tabular}{l@{~~~~~}| c ccc ccc ccc ccc ccc ccc  ccc}
\hline

 \multicolumn{1}{c|}{{} } &\multicolumn{3}{c|}{D1 $\rightarrow$ D2 } & \multicolumn{3}{c|}{D1 $\rightarrow$ D3 } & \multicolumn{3}{c|}{D2 $\rightarrow$ D1 } &\multicolumn{3}{c|}{D2 $\rightarrow$ D3 } & \multicolumn{3}{c|}{D3 $\rightarrow$ D1 } & \multicolumn{3}{c|}{D3 $\rightarrow$ D2 }  &  \multicolumn{3}{c}{\textit{Mean} }
 \\
   \multicolumn{1}{c|}{{}} & OS* & UNK & \multicolumn{1}{c|}{\textbf{\underline{HOS}}} &  OS* & UNK &  \multicolumn{1}{c|}{\textbf{\underline{HOS}}}  & OS* & UNK &  \multicolumn{1}{c|}{\textbf{\underline{HOS}}} & OS* & UNK & \multicolumn{1}{c|}{\textbf{\underline{HOS}}} &  OS* & UNK &  \multicolumn{1}{c|}{\textbf{\underline{HOS}}}  & OS* & UNK &  \multicolumn{1}{c|}{\textbf{\underline{HOS}}}  & OS* & UNK &  \multicolumn{1}{c}{\textbf{\underline{HOS}}} \\
\hline

 \multicolumn{1}{c|}{DeepAll}  & 37.61& 42.86 & \multicolumn{1}{c|}{40.06}  &37.54  & 35.21 & \multicolumn{1}{c|}{36.34}  & 35.66 & 44.12 & \multicolumn{1}{c|}{39.44}  &45.85 & 49.30 & \multicolumn{1}{c|}{47.51}  & 40.40 & 44.12 & \multicolumn{1}{c|}{42.18}  &  37.61& 32.14 & \multicolumn{1}{c|}{34.66}  &  39.11& 41.29 & \multicolumn{1}{c}{40.03}  \\
 
 \multicolumn{1}{c|}{{MM-SADA~\cite{Munro_2020_CVPR}} } & 36.89&  44.64& \multicolumn{1}{c|}{40.40}  & 36.99 &52.11  & \multicolumn{1}{c|}{43.27}  & 32.17 & 44.12 & \multicolumn{1}{c|}{37.21}  & 45.29& 50.70 & \multicolumn{1}{c|}{47.85}  & 34.16 &  41.18& \multicolumn{1}{c|}{37.34}  & 35.59 & 48.21 & \multicolumn{1}{c|}{40.95}  &  36.85&  46.83& \multicolumn{1}{c}{41.17}  \\
 
 \multicolumn{1}{c|}{{RNA-Net~\cite{Planamente_2022_WACV}}} & 42.51&  37.50& \multicolumn{1}{c|}{39.85}  & 30.56 &49.30  & \multicolumn{1}{c|}{37.73}  & 32.17 & 41.18 & \multicolumn{1}{c|}{36.12}  & 46.95& 33.80 & \multicolumn{1}{c|}{39.31}  & 41.65 & 35.29 & \multicolumn{1}{c|}{38.21}  & 48.56 &35.71  & \multicolumn{1}{c|}{41.16}  & 40.40 & 38.80 & \multicolumn{1}{c}{38.73}  \\

 \multicolumn{1}{c|}{{SimMMDG~\cite{dong2023SimMMDG}} } &33.86 & 41.07 & \multicolumn{1}{c|}{37.12}  & 39.31 & 56.34 & \multicolumn{1}{c|}{46.31}  &  37.16& 58.82 & \multicolumn{1}{c|}{45.54}  &40.31 & 71.83 & \multicolumn{1}{c|}{51.64}  &  31.17& 52.94 & \multicolumn{1}{c|}{39.24}  & 42.51 & 57.14 & \multicolumn{1}{c|}{48.75}  & 37.39 & 56.36 & \multicolumn{1}{c}{44.77}  \\

 \multicolumn{1}{c|}{{MOOSA (ours)}} &  35.73& 67.86 & \multicolumn{1}{c|}{\textbf{46.82}}  &  45.07  & 50.70 & \multicolumn{1}{c|}{\textbf{47.72}} & 40.15   & 64.71 & \multicolumn{1}{c|}{\textbf{49.55}}  & 46.62  & 59.15 & \multicolumn{1}{c|}{\textbf{52.15}}  &  38.15  &  55.88& \multicolumn{1}{c|}{\textbf{45.35}}  & 45.39 & 60.71 & \multicolumn{1}{c|}{\textbf{51.94}} & 41.85 & 59.84 & \multicolumn{1}{c}{\textbf{48.92}}  \\

\hline

\end{tabular} 
}
\vspace{0.1cm}
\caption{Multimodal \textbf{Open-set DA} with video and audio modalities on EPIC-Kitchens dataset.}
\vspace{-0.7cm}
\label{tab:epic-mmda} 
\end{table}

\noindent\textbf{{Multimodal Closed-set DG.}}
We also evaluate \textit{MOOSA} under the multimodal closed-set DG setup~\cite{dong2023SimMMDG}, where the label spaces of the source and target domains are identical, as presented in~\cref{tab:epic-dg2-less} and detailed further in~\cref{tab:epic-dg2}. In comparison with the current SOTA multimodal DG baseline, SimMMDG~\cite{dong2023SimMMDG}, our method achieves further improvements of up to $1.30\%$. This result suggests that the proposed multimodal self-supervised pretext tasks and entropy weighting mechanism contribute significantly to the model's ability to learn more discriminative features for enhanced generalization.

\begin{table*}[t!]
\centering
\resizebox{0.95\linewidth}{!}{
\begin{threeparttable}
\begin{tabular}{lccccccccccccc}
\toprule
& \multicolumn{3}{c}{\textbf{Modality}} & \multicolumn{5}{c}{\textbf{EPIC-Kitchens dataset}}& \multicolumn{4}{c}{\textbf{HAC dataset}}\\
\cmidrule(lr){2-4} \cmidrule(lr){5-9} \cmidrule(lr){10-13} 
\textbf{Method} & Video & Audio & Flow & D2, D3 $\rightarrow$ D1 & D1, D3 $\rightarrow$ D2 & D1, D2 $\rightarrow$ D3  & \textit{Mean}& & A, C $\rightarrow$ H & H, C $\rightarrow$ A & H, A $\rightarrow$ C  & \textit{Mean}\\

\midrule

DeepAll & $\checkmark$& $\checkmark$& $\checkmark$ & 55.63 & 59.20 & 58.01 & 57.61& &69.07&  71.30 &51.47 & 63.95  \\
MM-SADA~\cite{Munro_2020_CVPR} & $\checkmark$&$\checkmark$ &$\checkmark$ & 51.72  &  58.40 &  59.34  &   56.49&& 72.53 &  72.19 & 55.51&66.74   \\
RNA-Net~\cite{Planamente_2022_WACV} & $\checkmark$& $\checkmark$& $\checkmark$ & 52.41 & 57.20 & 60.16 & 56.59 &&  69.00&  73.40 & 51.65& 64.68 \\
SimMMDG~\cite{dong2023SimMMDG} & $\checkmark$& $\checkmark$& $\checkmark$ & \textbf{63.68} & 70.13 & 67.76 & 67.19& & 77.65 & \textbf{79.03} &56.62 & 71.10 \\
MOOSA (ours)& $\checkmark$& $\checkmark$& $\checkmark$  & 	63.22	 &  \textbf{70.27}& \textbf{68.38} & \textbf{67.29} & &	\textbf{78.01}	 & 77.15 &  \textbf{61.40} & \textbf{72.19} \\
\bottomrule
\end{tabular}

\end{threeparttable}
}
\vspace{0.1cm}
\caption{Multimodal \textbf{Closed-set DG} with all three modalities on EPIC-Kitchens and HAC datasets. The results for different combinations of any two modalities are in the Appendix.}
\vspace{-0.7cm}
\label{tab:epic-dg2-less}
\end{table*}



\subsection{Ablation Studies and Analysis}

\noindent\textbf{{Disparate label sets across domains.}} In previous MM-OSDG experiments, we assumed that the label sets of all source domains were identical, and the target domain included all label sets from source domains along with additional unknown classes. However, in practical scenarios, different source domains may comprise distinct label sets, while the target domain might not include all label sets present in the source domains. \cref{tab:epic-labelsets} illustrates the results under this more challenging setup, where our method consistently demonstrates improvements in comparison to other baselines. The details of the label split under this setup are provided in \cref{tab:diff-s} and \cref{fig:splits}.

\begin{table}[t!]
\centering
\resizebox{0.75\textwidth}{!}{
\begin{tabular}{l@{~~~~~}| c ccc ccc ccc ccc}
\hline

 \multicolumn{1}{c|}{{} } &\multicolumn{3}{c|}{D2, D3 $\rightarrow$ D1 } & \multicolumn{3}{c|}{D1, D3 $\rightarrow$ D2 } & \multicolumn{3}{c|}{D1, D2 $\rightarrow$ D3 }  &  \multicolumn{3}{c}{\textit{Mean} }
 \\
   \multicolumn{1}{c|}{{}} & OS* & UNK & \multicolumn{1}{c|}{\textbf{\underline{HOS}}} &  OS* & UNK &  \multicolumn{1}{c|}{\textbf{\underline{HOS}}}  & OS* & UNK &  \multicolumn{1}{c|}{\textbf{\underline{HOS}}}  & OS* & UNK &  \multicolumn{1}{c}{\textbf{\underline{HOS}}} \\
\hline

 \multicolumn{1}{c|}{DeepAll} & 49.04 & 35.29 & \multicolumn{1}{c|}{41.05}  & 52.47 & 30.36 & \multicolumn{1}{c|}{38.46}  &  50.45& 39.44 & \multicolumn{1}{c|}{44.27}  & 50.65 & 35.03 & \multicolumn{1}{c}{41.26}  \\

 \multicolumn{1}{c|}{{MM-SADA~\cite{Munro_2020_CVPR}} } & 41.76&  47.06& \multicolumn{1}{c|}{44.25}  & 52.47 & 39.29 & \multicolumn{1}{c|}{44.93}  & 50.27 & 38.03 & \multicolumn{1}{c|}{43.30}  & 48.17 & 41.46 & \multicolumn{1}{c}{44.16}  \\
 
 \multicolumn{1}{c|}{{RNA-Net~\cite{Planamente_2022_WACV}}} & 48.28& 38.24 & \multicolumn{1}{c|}{42.67}  &  51.61& 32.14 & \multicolumn{1}{c|}{39.61}  & 48.28 & 35.21 & \multicolumn{1}{c|}{40.72}  &  49.39& 35.20 & \multicolumn{1}{c}{41.00}  \\

 \multicolumn{1}{c|}{{SimMMDG~\cite{dong2023SimMMDG}} } & 48.28& 47.06 & \multicolumn{1}{c|}{47.66}  & 39.35 & 62.50 & \multicolumn{1}{c|}{48.30}  &43.92  &54.93  & \multicolumn{1}{c|}{48.81}  & 43.85 & 54.83 & \multicolumn{1}{c}{48.26}  \\

 \multicolumn{1}{c|}{{MOOSA (ours)}} & 49.04 & 61.76 & \multicolumn{1}{c|}{\textbf{54.67}}  & 46.24 & 57.14 & \multicolumn{1}{c|}{\textbf{51.11}}  &  45.92& 59.15 & \multicolumn{1}{c|}{\textbf{51.70}}  & 47.07 & 59.35 & \multicolumn{1}{c}{\textbf{52.49}}  \\

\hline

\end{tabular} 
}
\vspace{0.1cm}
\caption{Abation on disparate label sets  across domains for MM-OSDG with video and audio modalities on EPIC-Kitchens dataset. }
\vspace{-0.7cm}
\label{tab:epic-labelsets} 
\end{table}

\noindent\textbf{{Compared with unimodal open-set DG:}} \cref{tab:epic-unimodal} offers a comparison against unimodal open-set DG algorithms that exclusively utilize either video, audio, or optical flow as inputs. We selected the SOTA unimodal open-set DG method, MEDIC, as our baseline. By leveraging information from multiple modalities, our \textit{MOOSA} demonstrates substantial improvements up to $7.78\%$ in  $HOS$ compared to unimodal open-set DG methods.

\begin{table}[t!]
\centering
\resizebox{0.85\textwidth}{!}{
\begin{tabular}{l@{~~~~~}| c ccc ccc ccc ccc}
\hline

 \multicolumn{1}{c|}{{} } &\multicolumn{3}{c|}{D2, D3 $\rightarrow$ D1 } & \multicolumn{3}{c|}{D1, D3 $\rightarrow$ D2 } & \multicolumn{3}{c|}{D1, D2 $\rightarrow$ D3 }  &  \multicolumn{3}{c}{\textit{Mean} }
 \\
   \multicolumn{1}{c|}{{}} & OS* & UNK & \multicolumn{1}{c|}{\textbf{\underline{HOS}}} &  OS* & UNK &  \multicolumn{1}{c|}{\textbf{\underline{HOS}}}  & OS* & UNK &  \multicolumn{1}{c|}{\textbf{\underline{HOS}}}  & OS* & UNK &  \multicolumn{1}{c}{\textbf{\underline{HOS}}} \\
\hline

 \multicolumn{1}{c|}{{MEDIC~\cite{wang2023generalizable} (V)} } & 36.41& 58.82 & \multicolumn{1}{c|}{44.98}  &  34.73& 66.07 & \multicolumn{1}{c|}{45.53}  &  31.34& 59.15 & \multicolumn{1}{c|}{40.97}  & 34.16 & 61.35 & \multicolumn{1}{c}{43.83}  \\

 \multicolumn{1}{c|}{{MEDIC~\cite{wang2023generalizable} (A)} } &  22.44& 47.06 & \multicolumn{1}{c|}{30.39}  &  27.67&  62.50& \multicolumn{1}{c|}{38.35}  &  35.44& 46.48 & \multicolumn{1}{c|}{40.21}  & 28.52 &52.01  & \multicolumn{1}{c}{36.32}  \\
 
 \multicolumn{1}{c|}{{MEDIC~\cite{wang2023generalizable} (F)} } & 36.91 & 64.71 & \multicolumn{1}{c|}{47.00}  & 45.97 & 58.93 & \multicolumn{1}{c|}{51.65}  & 42.30 & 56.34 & \multicolumn{1}{c|}{48.32}  & 41.73 & 59.99 & \multicolumn{1}{c}{48.99}  \\
 
 \multicolumn{1}{c|}{{MOOSA (ours) (V+A+F)}} & 46.13 &73.52  & \multicolumn{1}{c|}{\textbf{56.70}}  &53.03  &  67.86& \multicolumn{1}{c|}{\textbf{59.53}}  & 48.84 & 60.56 & \multicolumn{1}{c|}{\textbf{54.07}}  & 49.33 & 67.31 & \multicolumn{1}{c}{\textbf{56.77}}  \\

\hline

\end{tabular} 
}
\vspace{0.1cm}
\caption{Compared with  unimodal open-set DG. V: video, A: audio, F: optical flow.}
\vspace{-0.7cm}
\label{tab:epic-unimodal} 
\end{table}

\noindent\textbf{{Ablation on each proposed module.}} We conducted comprehensive ablation studies to investigate the role of each module of our method on the EPIC-Kitchens dataset, as illustrated in~\cref{tab:module}. The inclusion of either masked cross-modal translation or multimodal Jigsaw puzzles as pretext tasks demonstrates performance benefits. The combination of these two tasks further enhances the results, verifying their complementarity. The best performance is achieved when incorporating entropy weighting and minimization to balance the loss from different modalities.

\setlength{\tabcolsep}{2pt}
\begin{table}[t!]
\begin{minipage}[t]{0.5\columnwidth}
\centering
\begin{adjustbox}{width=0.95\linewidth}

\begin{tabular}{l@{~~~~~}cccc  c}
\hline

         MulJig & MaskedTrans  & EntWei & \multicolumn{1}{c|}{EntMin}  &  \multicolumn{1}{c}{\textit{Mean} \textbf{\underline{HOS}}} \\
\hline

   & & &  \multicolumn{1}{c|}{} &  \multicolumn{1}{c}{50.62}  \\

   $\checkmark$& & &  \multicolumn{1}{c|}{} &  \multicolumn{1}{c}{51.88}  \\

   & $\checkmark$& &  \multicolumn{1}{c|}{} &  \multicolumn{1}{c}{51.60}  \\

  $\checkmark$  & $\checkmark$& &  \multicolumn{1}{c|}{} &  \multicolumn{1}{c}{53.91}  \\

  $\checkmark$  & $\checkmark$& $\checkmark$&  \multicolumn{1}{c|}{} & \multicolumn{1}{c}{54.27}  \\

    $\checkmark$ & $\checkmark$& $\checkmark$&  \multicolumn{1}{c|}{$\checkmark$} &  \multicolumn{1}{c}{\textbf{54.57}}  \\

\hline
\end{tabular} 

\end{adjustbox}
\vspace{0.1cm}
\caption{Ablations of each proposed module in our \textit{MOOSA}.}

\label{tab:module}
\end{minipage}
\hfill
\begin{minipage}[t]{0.5\columnwidth}
\centering

\begin{adjustbox}{width=0.8\linewidth}
\begin{tabular}{l@{~~~~~}| c c}
\hline

   \multicolumn{1}{c|}{{Self-supervised Tasks}} & \multicolumn{1}{c}{\textit{Mean} \textbf{\underline{HOS}}} \\
\hline

 
 \multicolumn{1}{c|}{Multimodal Alignment~\cite{Munro_2020_CVPR}} & \multicolumn{1}{c}{51.17}  \\
 
  \multicolumn{1}{c|}{Cross-modal Translation~\cite{dong2023SimMMDG}} & \multicolumn{1}{c}{48.94}  \\

  \multicolumn{1}{c|}{Relative Norm Alignment~\cite{Planamente_2022_WACV}} & \multicolumn{1}{c}{50.55}  \\
  
  \multicolumn{1}{c|}{Cross-modal Distillation~\cite{gupta2016cross}} & {50.88}  \\
 
 \multicolumn{1}{c|}{Masked Cross-modal Translation} &  \multicolumn{1}{c}{51.60}  \\

 \multicolumn{1}{c|}{Multimodal Jigsaw Puzzles} &   \multicolumn{1}{c}{51.88}  \\

 \multicolumn{1}{c|}{MaskedTrans+MulJig} &   \multicolumn{1}{c}{\textbf{53.91}}  \\

\hline

\end{tabular} 
\end{adjustbox}
\vspace{0.1cm}
\caption{Abation on different self-supervised tasks.}

\label{tab:sst}
\end{minipage}
\vspace{-0.8cm}
\end{table}

\noindent\textbf{{On the effectiveness of other self-supervised pretext tasks.}} We also explore the feasibility of replacing our multimodal self-supervised tasks with alternative pretext tasks. As depicted in~\cref{tab:sst}, multimodal alignment~\cite{Munro_2020_CVPR} and cross-modal distillation~\cite{gupta2016cross} prove to be effective as pretext tasks for MM-OSDG. However, our proposed masked cross-modal translation and multimodal Jigsaw puzzles emerge as superior self-supervised tasks in this context.



\begin{figure}[t!]
\begin{minipage}[t]{0.45\columnwidth}
  \centering  \includegraphics[width=\linewidth]{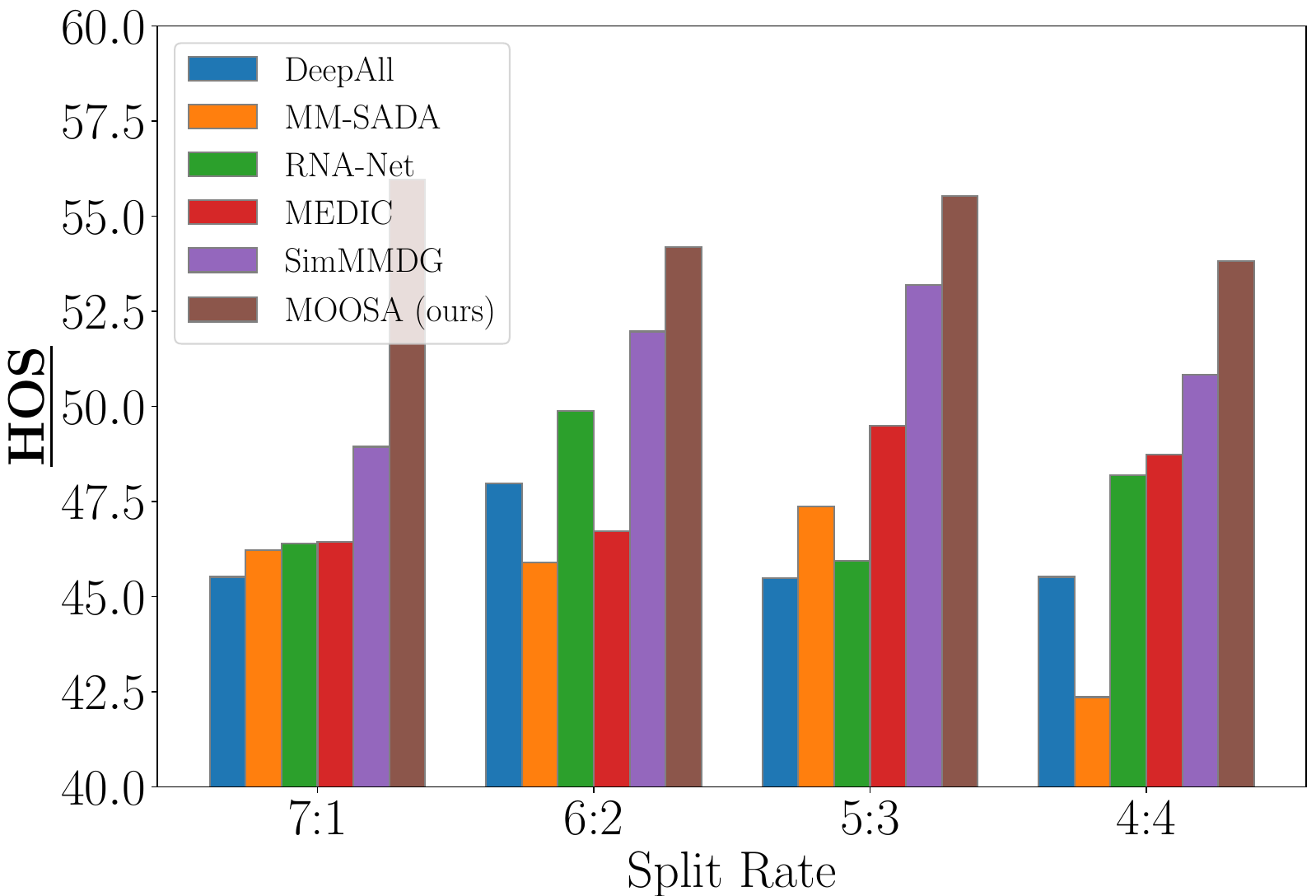}
   \vspace{-0.6cm}
   \caption{The average $HOS$ with the varying known-unknown split rates on EPIC-Kitchens dataset.}
   \label{fig:split}
\end{minipage}
\begin{minipage}[t]{0.5\columnwidth}
\centering  \includegraphics[width=\linewidth]{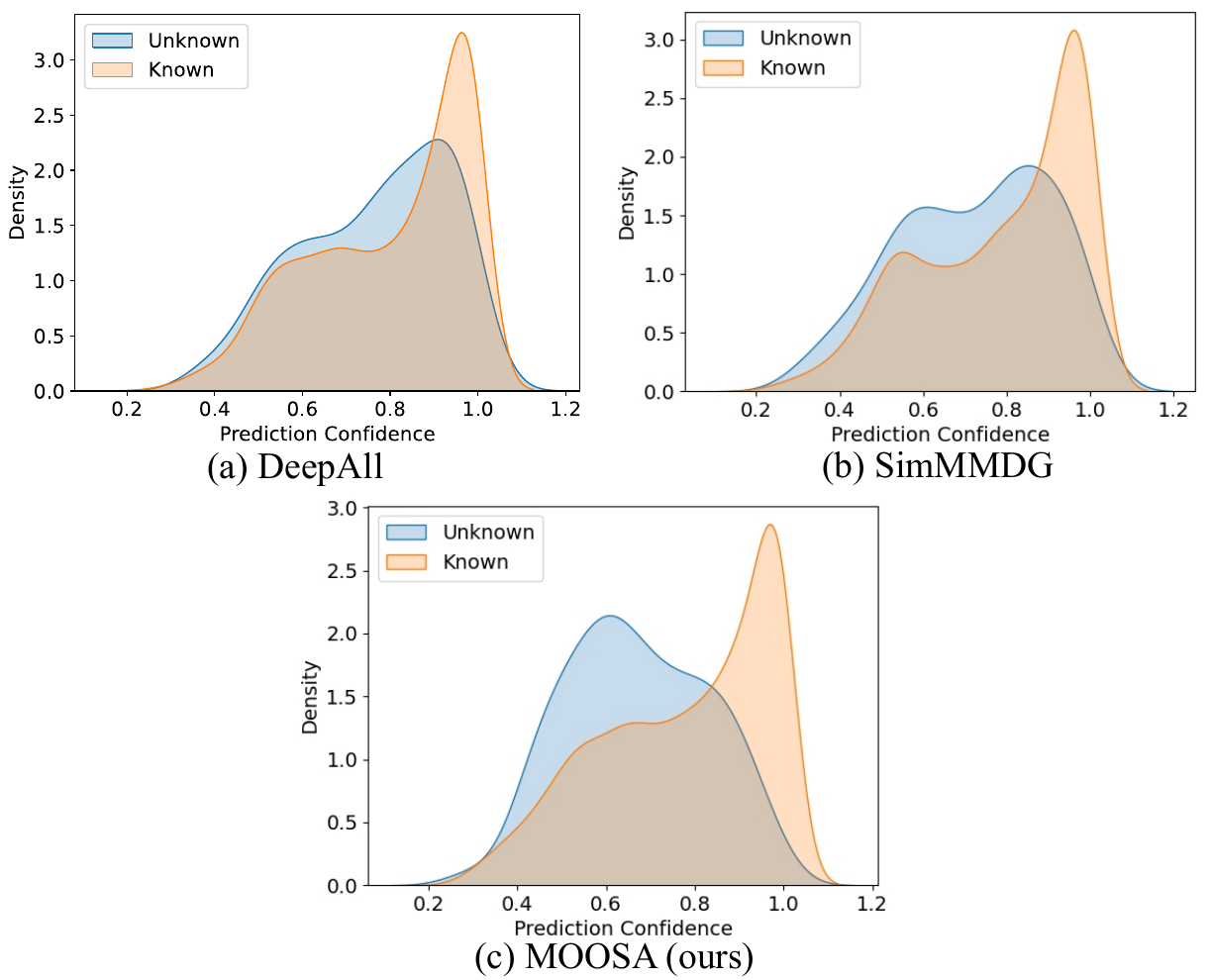}
   \vspace{-0.6cm}
   \caption{The distribution curves of prediction confidence for unknown and known classes obtained using various methods.}
   \label{fig:score}
\end{minipage}
   \vspace{-0.35cm}
\end{figure}

\noindent\textbf{{Varying the ratio of known to unknown classes.}} The term "openness" denotes the proportion of unknown target classes during the test phase. To assess the adaptability of our model across different levels of openness, we perform experiments on the EPIC-Kitchens dataset. The results presented in~\cref{fig:split} demonstrate that increasing the number of open classes presents more challenges for the classification task, leading to reduced model accuracy. However, our method consistently outperforms alternative approaches, showcasing significant improvements and validating the robustness of our model across diverse scenarios.

\noindent\textbf{{Visualization of confidence score distributions for known and unknown classes.}}
In~\cref{fig:score}, we illustrate the distribution curves of prediction confidence for unknown and known classes across different methods. A smaller overlap between unknown and known data indicates superior open-class detection performance, whereas a larger overlap suggests weaker detection results. The unknown curve (depicted in blue) for our method exhibits a distinctive peak at a lower position, resulting in minimal overlap with other known data. This distinct characteristic underscores the superior open-class detection capability achieved by our method. This success can be attributed to the high-quality in-distribution feature representation.


\section{Conclusion}
In this paper, we introduce the Multimodal Open-Set Domain
Generalization problem and propose a novel solution \textit{MOOSA} employing self-supervised pretext tasks. Our approach involves the development of two complementary multimodal self-supervised tasks, namely Masked Cross-modal Translation and Multimodal Jigsaw Puzzles, aimed at learning multimodal representative features for robust generalization and effective open-class detection. To maintain a balance in the loss across different modalities, we introduce an Entropy Weighting mechanism during training. Furthermore, we extend our framework to address Multimodal Open-Set Domain Adaptation. Through extensive evaluations on the EPIC-Kitchens and HAC datasets, our method demonstrates superior performance compared to competing baselines. Notably, our approach exhibits robustness across variations in label sets, openness levels, and parameter choices. In future work, methods proposed in Multimodal OOD Detection~\cite{dong2024multiood} could be good alternatives for open-class detection.

\section*{Acknowledgments}

The authors acknowledge the support of "In-service diagnostics of the catenary/pantograph and wheelset axle systems through intelligent algorithms" (SENTINEL) project, supported by the ETH Mobility Initiative.

%
%
\bibliographystyle{splncs04}
\bibliography{egbib}

\appendix

\section{Details on Dataset Splits}
For each dataset, we show the exact class splits for each
domain.

\noindent\textbf{EPIC-Kitchens dataset}~\cite{Munro_2020_CVPR} consists of three domains, denoted as D1, D2, and D3, corresponding to actions recorded in three different kitchen styles. This dataset comprises eight actions, namely, ‘put’, ‘take’, ‘open’, ‘close’, ‘wash’, ‘cut’, ‘mix’, and ‘pour’. All three domains share a common label set of eight classes.  We organize the action names in alphabetic order and assign an index
to each category, 0-close, 1-cut, 2-mix, 3-open, 4-pour, 5-put, 6-take, 7-wash. Each domain is employed as the target domain, while the other two domains function as source domains, forming three distinct cross-domain tasks  (D2, D3 $\rightarrow$ D1, and D1, D3 $\rightarrow$ D2, and D1, D2 $\rightarrow$ D3). To construct the open-set situations, we designate the first class as unknown and the remaining classes as known. The specific categories contained in each domain are shown in \cref{tab:epic-s}. 

\noindent\textbf{HAC dataset}~\cite{dong2023SimMMDG}  comprises seven actions (‘sleeping’, ‘watching tv’, ‘eating’, ‘drinking’, ‘swimming’, ‘running’, and ‘opening door’) executed by humans, animals, and cartoon figures, forming three distinct domains H, A, and C. The three domains have the same label set of 7 classes. We organize the action names in alphabetic order and assign an index
to each category, 0-drinking, 1-eating, 2-opening door, 3-running, 4-sleeping, 5-swimming, 6-watching tv. We use each domain as the target domain and the other two domains as source domains
to form three cross-domain tasks (A, C $\rightarrow$ H, and H, C $\rightarrow$ A, and H, A $\rightarrow$ C).  In the creation of open-set scenarios, the first class is designated as unknown, whereas the remaining classes are considered known. The specific categories contained in each domain are shown in \cref{tab:hac-s}.

\setlength{\tabcolsep}{2pt}
\begin{table}[ht!]
\begin{minipage}[t]{0.45\columnwidth}
\centering
\begin{adjustbox}{width=0.7\linewidth}

\begin{tabular}{ccccc}
\hline Domain & & & & Classes \\
\hline Source-1 & & & & $1,2,3,4,5,6,7$ \\
Source-2 & & & & $1,2,3,4,5,6,7$ \\
Target & & & & $0,1,2,3,4,5,6,7$ \\
\hline
\end{tabular} 

\end{adjustbox}
\vspace{0.1cm}
\caption{Open-set class splits of EPIC-Kitchens dataset.}

\label{tab:epic-s}
\end{minipage}
\hfill
\begin{minipage}[t]{0.45\columnwidth}
\centering

\begin{adjustbox}{width=0.7\linewidth}
\begin{tabular}{ccccc}
\hline Domain & & & & Classes \\
\hline Source-1 & & & & $1,2,3,4,5,6$ \\
Source-2 &  & & &$1,2,3,4,5,6$ \\
Target & & & & $0,1,2,3,4,5,6$ \\
\hline
\end{tabular} 
\end{adjustbox}
\vspace{0.1cm}
\caption{Open-set class splits of HAC dataset.}

\label{tab:hac-s}
\end{minipage}
\end{table}

\section{Implementation Details}
\label{imple}
Our experiments encompass three modalities: video, audio, and optical flow. We adopt SimMMDG~\cite{dong2023SimMMDG} as the base framework by splitting the features within each modality into modality-specific and modality-shared components due to its demonstrated efficacy in multimodal DG tasks. In line with~\cite{dong2023SimMMDG}, we employ the MMAction2~\cite{2020mmaction2} toolkit for experiments. For encoding visual information, we utilize the SlowFast network~\cite{Feichtenhofer_2019_ICCV} initialized with Kinetics-400~\cite{kay2017kinetics} pre-trained weights. The audio encoder employs ResNet-18~\cite{7780459}, initialized with weights from the VGGSound pre-trained checkpoint~\cite{9053174}. Similarly, the optical flow encoder employs the SlowFast network with a slow-only pathway, initialized with Kinetics-400~\cite{kay2017kinetics} pre-trained weights. The dimensions of the unimodal embedding $\mathbf{E}$ for video, audio, and optical flow are $2304$, $512$, and $2048$, respectively.
 We employ a multi-layer perceptron with two hidden layers of size $2048$ to instantiate the masked cross-modal translation $MLP_{\mathbf{E}^{i} \rightarrow \mathbf{E}^{j}}$ and select a mask ratio $\beta$ of $0.7$. For multimodal Jigsaw puzzles, we set $P$ to $4$ and $Q$ to $128$. The classifier $h_p$ is an MLP with two layers and hidden size $512$. The Adam optimizer~\cite{Adam} is utilized with a learning rate of $0.0001$ and a batch size of $16$. The scalar temperature parameter $T$ in entropy weighting is set to $1.0$. Additionally, we set $\alpha_{1} = 0.1$, $\alpha_{2} = 1.0$, and $\alpha_{3} = 0.1$. 
Finally, the network is trained for $20$ epochs on an RTX 3090 GPU, requiring approximately $20$ hours. The model with the best performance on the validation dataset is selected.

\section{More Experimental Results}
\noindent\textbf{{Multimodal Open-set DG.}}
\begin{table}[t!]
\centering
\resizebox{\textwidth}{!}{
\begin{tabular}{l@{~~~~~}cc ccc ccc ccc ccc ccc}
\hline

 & &\multicolumn{3}{c|}{\textbf{Modality} } &\multicolumn{3}{c|}{A, C $\rightarrow$ H } & \multicolumn{3}{c|}{H, C $\rightarrow$ A } & \multicolumn{3}{c|}{H, A $\rightarrow$ C }  &  \multicolumn{3}{c}{\textit{Mean} }
 \\
        &  & Video  & Audio & \multicolumn{1}{c|}{Flow} & OS* & UNK & \multicolumn{1}{c|}{\textbf{\underline{HOS}}} &  OS* & UNK &  \multicolumn{1}{c|}{\textbf{\underline{HOS}}}  & OS* & UNK &  \multicolumn{1}{c|}{\textbf{\underline{HOS}}}  & OS* & UNK &  \multicolumn{1}{c}{\textbf{\underline{HOS}}} \\
\hline

{DeepAll} &  & $\checkmark$& $\checkmark$&  \multicolumn{1}{c|}{} & 62.83 & 60.10 & \multicolumn{1}{c|}{61.43}  & 67.40 & 51.13 & \multicolumn{1}{c|}{58.15}  & 40.23 & 55.56 & \multicolumn{1}{c|}{46.67}  & 56.82 & 55.60 & \multicolumn{1}{c}{55.42}  \\

{MM-SADA~\cite{Munro_2020_CVPR}} &  & $\checkmark$& $\checkmark$&  \multicolumn{1}{c|}{} & 62.32 & 77.78 & \multicolumn{1}{c|}{69.20}  & 68.95 & 61.65 & \multicolumn{1}{c|}{65.10}  & 33.37 & 50.79 & \multicolumn{1}{c|}{40.28}  & 54.88 & 63.41 & \multicolumn{1}{c}{58.19}  \\

{RNA-Net~\cite{Planamente_2022_WACV}} &  & $\checkmark$& $\checkmark$&  \multicolumn{1}{c|}{} &59.29  & 74.24 & \multicolumn{1}{c|}{65.93}  & 72.19 & 58.65 & \multicolumn{1}{c|}{64.72}  & 32.33 & 67.46 & \multicolumn{1}{c|}{43.71}  & 54.60 & 66.78 & \multicolumn{1}{c}{58.12}  \\

{MEDIC~\cite{wang2023generalizable}} &  &$\checkmark$&  $\checkmark$&  \multicolumn{1}{c|}{} & 51.05 & 80.30 & \multicolumn{1}{c|}{62.42}  & 53.69&  75.19& \multicolumn{1}{c|}{62.64}   &  35.76&  66.67& \multicolumn{1}{c|}{46.55}   & 46.83 & 74.05 & \multicolumn{1}{c}{57.20}  \\

{SimMMDG~\cite{dong2023SimMMDG}} &  & $\checkmark$& $\checkmark$&  \multicolumn{1}{c|}{} & 68.88 & 80.81 & \multicolumn{1}{c|}{\textbf{74.37}}   & 59.38 & 75.94 & \multicolumn{1}{c|}{66.65}  & 35.97 & 68.25 & \multicolumn{1}{c|}{47.11}  & 54.74 & 75.00 & \multicolumn{1}{c}{62.71}  \\

{MOOSA (ours) } &  & $\checkmark$& $\checkmark$&  \multicolumn{1}{c|}{} & 70.40 &  77.78 & \multicolumn{1}{c|}{73.90}  & 66.62 & 71.43 & \multicolumn{1}{c|}{\textbf{68.94}}  & 36.80 & 76.19 & \multicolumn{1}{c|}{\textbf{49.63}}  & 57.94 & 75.13 & \multicolumn{1}{c}{\textbf{64.16}}  \\

\hline

{DeepAll} &  & $\checkmark$& &  \multicolumn{1}{c|}{$\checkmark$} & 69.05 & 58.59 & \multicolumn{1}{c|}{63.39}  & 70.89 & 45.11 & \multicolumn{1}{c|}{55.14}  & 40.12 & 64.29 & \multicolumn{1}{c|}{49.41}  &  60.02& 56.00 & \multicolumn{1}{c}{55.98}  \\

{MM-SADA~\cite{Munro_2020_CVPR}} &  & $\checkmark$& &  \multicolumn{1}{c|}{$\checkmark$} & 64.17 & 65.15 & \multicolumn{1}{c|}{64.66}  & 65.59 & 62.41 & \multicolumn{1}{c|}{63.96}  & 38.36 & 58.73 & \multicolumn{1}{c|}{46.41}  & 56.04 & 62.10 & \multicolumn{1}{c}{58.34}  \\

{RNA-Net~\cite{Planamente_2022_WACV}} &  & $\checkmark$& &  \multicolumn{1}{c|}{$\checkmark$} & 74.26 & 53.54 & \multicolumn{1}{c|}{62.22}  & 68.08 & 58.65 & \multicolumn{1}{c|}{63.44}  & 39.71 & 53.97 & \multicolumn{1}{c|}{45.75}  & 60.68 & 55.39 & \multicolumn{1}{c}{57.14}  \\

{MEDIC~\cite{wang2023generalizable}} &  &$\checkmark$&  &  \multicolumn{1}{c|}{$\checkmark$} &58.45  & 76.26 & \multicolumn{1}{c|}{66.18}  & 53.17& 69.17 & \multicolumn{1}{c|}{60.12}   & 37.01 & 66.67 & \multicolumn{1}{c|}{47.59}   & 49.54 & 70.70 & \multicolumn{1}{c}{57.96}  \\

{SimMMDG~\cite{dong2023SimMMDG}} &  & $\checkmark$& &  \multicolumn{1}{c|}{$\checkmark$} & 74.43 & 59.09 & \multicolumn{1}{c|}{65.88}  & 61.32 & 72.93 & \multicolumn{1}{c|}{66.62}  & 37.42 & 67.46 & \multicolumn{1}{c|}{48.14}  & 57.72 & 66.49 & \multicolumn{1}{c}{60.21}  \\

{MOOSA (ours)} &  & $\checkmark$& &  \multicolumn{1}{c|}{$\checkmark$} & 70.98 & 68.69 & \multicolumn{1}{c|}{\textbf{69.82}}  &65.59  & 69.92 & \multicolumn{1}{c|}{\textbf{67.69}}  & 45.84 & 63.49 & \multicolumn{1}{c|}{\textbf{53.24}}  &  60.80& 67.37 & \multicolumn{1}{c}{\textbf{63.58}}  \\

\hline

{DeepAll} &  & &$\checkmark$ &  \multicolumn{1}{c|}{$\checkmark$} & 47.94 & 64.65 & \multicolumn{1}{c|}{55.05}  &  60.67& 33.08 & \multicolumn{1}{c|}{42.82}  & 33.37 & 50.00 & \multicolumn{1}{c|}{40.02}  & 47.33 & 49.24 & \multicolumn{1}{c}{45.96}  \\

{MM-SADA~\cite{Munro_2020_CVPR}} &  & &$\checkmark$ &  \multicolumn{1}{c|}{$\checkmark$} & 43.48 & 52.53 & \multicolumn{1}{c|}{47.58}  & 44.24 & 53.38 & \multicolumn{1}{c|}{48.39}  & 35.34 & 50.00 & \multicolumn{1}{c|}{41.41}  &  41.02& 51.97 & \multicolumn{1}{c}{45.79}  \\

{RNA-Net~\cite{Planamente_2022_WACV}} &  &&  $\checkmark$&  \multicolumn{1}{c|}{$\checkmark$} & 47.43 & 58.08 & \multicolumn{1}{c|}{52.22}  & 56.79 & 42.86 & \multicolumn{1}{c|}{48.85}  & 34.82 & 57.14 & \multicolumn{1}{c|}{43.27}  &46.35  & 52.69 & \multicolumn{1}{c}{48.11}  \\

{MEDIC~\cite{wang2023generalizable}} &  && $\checkmark$ &  \multicolumn{1}{c|}{$\checkmark$} & 45.08 &  66.67& \multicolumn{1}{c|}{53.79}  & 47.61&  60.90& \multicolumn{1}{c|}{53.44}   &  34.62&60.32  & \multicolumn{1}{c|}{43.99}   &  42.44& 62.63 & \multicolumn{1}{c}{50.41}  \\

{SimMMDG~\cite{dong2023SimMMDG}} &  & & $\checkmark$&  \multicolumn{1}{c|}{$\checkmark$} & 46.93 & 69.70 & \multicolumn{1}{c|}{56.09}  & 45.67 & 60.15 & \multicolumn{1}{c|}{51.92}  & 36.28 & 66.67 & \multicolumn{1}{c|}{\textbf{46.99}} & 42.96 & 65.51 & \multicolumn{1}{c}{51.67}  \\

{MOOSA (ours)} &  & &$\checkmark$ &  \multicolumn{1}{c|}{$\checkmark$} & 45.75 & 76.26 & \multicolumn{1}{c|}{\textbf{57.19}}  &51.36 &   59.40 & \multicolumn{1}{c|}{\textbf{55.09}}  &36.38 & 64.29 & \multicolumn{1}{c|}{46.47}  & 44.50 & 66.65 & \multicolumn{1}{c}{\textbf{52.92}}  \\

\hline

{DeepAll} &  & $\checkmark$&$\checkmark$ &  \multicolumn{1}{c|}{$\checkmark$} & 66.02 &71.72  & \multicolumn{1}{c|}{68.75}  & 74.51 & 54.14 & \multicolumn{1}{c|}{62.71}  & 44.59 & 55.56 & \multicolumn{1}{c|}{49.48}  &  61.71& 60.47 & \multicolumn{1}{c}{60.31}  \\

{MM-SADA~\cite{Munro_2020_CVPR}} &  & $\checkmark$&$\checkmark$ &  \multicolumn{1}{c|}{$\checkmark$} & 60.13 & 72.22 & \multicolumn{1}{c|}{65.63}  & 76.46 & 52.63 & \multicolumn{1}{c|}{62.35}  &  40.64  & 56.35 & \multicolumn{1}{c|}{47.23}  &59.08  & 60.40 & \multicolumn{1}{c}{58.40}  \\

{RNA-Net~\cite{Planamente_2022_WACV}} &  & $\checkmark$&$\checkmark$ &  \multicolumn{1}{c|}{$\checkmark$} & 70.31 & 67.68 & \multicolumn{1}{c|}{68.97}  & 72.06 & 51.88 & \multicolumn{1}{c|}{60.33}  &46.99  &55.56 & \multicolumn{1}{c|}{50.91}  & 63.12 & 58.37 & \multicolumn{1}{c}{60.07}  \\

{MEDIC~\cite{wang2023generalizable}} &  &$\checkmark$& $\checkmark$ &  \multicolumn{1}{c|}{$\checkmark$} & 56.18 & 77.78 & \multicolumn{1}{c|}{65.24}  &58.47 & 69.17 & \multicolumn{1}{c|}{63.37}   & 42.93 & 61.11 & \multicolumn{1}{c|}{50.43}   &  52.53& 69.35 & \multicolumn{1}{c}{59.68}  \\

{SimMMDG~\cite{dong2023SimMMDG}} &  & $\checkmark$& $\checkmark$&  \multicolumn{1}{c|}{$\checkmark$} & 62.83 & 77.27 & \multicolumn{1}{c|}{69.30}  & 63.78 & 75.19 & \multicolumn{1}{c|}{69.01}  & 42.41 & 64.29 & \multicolumn{1}{c|}{51.11}  &56.34  & 72.25 & \multicolumn{1}{c}{63.14}  \\

{MOOSA (ours)} &  & $\checkmark$&$\checkmark$ &  \multicolumn{1}{c|}{$\checkmark$} & 66.69 & 78.79 & \multicolumn{1}{c|}{\textbf{72.24}}  &66.36 & 75.94 & \multicolumn{1}{c|}{\textbf{70.83}}  &  45.01& 62.70 & \multicolumn{1}{c|}{\textbf{52.40}}  & 59.35& 72.48 & \multicolumn{1}{c}{\textbf{65.16}}  \\

\hline
\end{tabular} 
}
\vspace{0.1cm}
\caption{Multimodal \textbf{Open-set DG} with different combinations of modalities on HAC dataset.}
\label{tab:hac-open} 
\end{table}
\cref{tab:hac-open} presents the results under the Multimodal Open-set DG setting on HAC dataset~\cite{dong2023SimMMDG} with different combinations of any two modalities, as well as all three modalities. Our method consistently outperforms all baselines by a significant margin in most cases, yielding average improvements in $HOS$ of up to $3.37\%$.

\noindent\textbf{{Multimodal Closed-set DG.}}
Detailed results under the Multimodal Closed-set DG setting with different combinations of any two modalities, as well as all three modalities, are presented in~\cref{tab:epic-dg2}. In comparison with the current SOTA multimodal DG baseline SimMMDG~\cite{dong2023SimMMDG}, our method attains further improvements up to $1.30\%$. This observation suggests that the proposed multimodal self-supervised pretext tasks and entropy weighting mechanism contribute to the model's ability to learn more discriminative features for enhanced generalization.

\begin{table*}[t!]
\centering
\resizebox{\linewidth}{!}{
\begin{threeparttable}
\begin{tabular}{lccccccccccccc}
\toprule
& \multicolumn{3}{c}{\textbf{Modality}} & \multicolumn{5}{c}{\textbf{EPIC-Kitchens dataset}}& \multicolumn{4}{c}{\textbf{HAC dataset}}\\
\cmidrule(lr){2-4} \cmidrule(lr){5-9} \cmidrule(lr){10-13} 
\textbf{Method} & Video & Audio & Flow & D2, D3 $\rightarrow$ D1 & D1, D3 $\rightarrow$ D2 & D1, D2 $\rightarrow$ D3  & \textit{Mean}& & A, C $\rightarrow$ H & H, C $\rightarrow$ A & H, A $\rightarrow$ C  & \textit{Mean}\\

\midrule

DeepAll & $\checkmark$& $\checkmark$&    & 47.13 & 55.73 & 57.17 &  53.34& &66.55  & 72.85  &  45.77 & 61.72 \\
MM-SADA~\cite{Munro_2020_CVPR} & $\checkmark$& $\checkmark$& &   49.20  & 60.40  & 59.14     & 56.25&  & 65.47  &  72.52 & 44.30  & 60.76 \\
RNA-Net~\cite{Planamente_2022_WACV} & $\checkmark$& $\checkmark$&  & 52.18 & 59.47 & 60.88  &57.51 & &60.20  & 73.95  &  48.90 &  61.02\\
SimMMDG~\cite{dong2023SimMMDG}& $\checkmark$& $\checkmark$&   & 		57.93 & 65.47 & 66.32 & 63.24& &		74.77 & 77.81 & \textbf{53.68} & 68.75 \\

MOOSA (ours)& $\checkmark$& $\checkmark$&   &	\textbf{59.31} &\textbf{68.00} & \textbf{64.37}  & \textbf{63.89} & &	\textbf{76.86}	 & \textbf{78.15} & 52.11 & \textbf{69.04} \\

\midrule

DeepAll & $\checkmark$& &$\checkmark$ & 55.17 & 62.93 & 60.37 &59.49& & 76.78  &  70.64 &49.63 &  65.68 \\
MM-SADA~\cite{Munro_2020_CVPR} & $\checkmark$& &$\checkmark$ & 47.13  & 57.60  &  59.34  & 54.69 & & 69.79 & 69.76  &49.45 & 63.00 \\
RNA-Net~\cite{Planamente_2022_WACV} & $\checkmark$& & $\checkmark$& 54.71 & 61.87 & 58.21 &58.26& &  77.14 &74.94  &42.00 &  64.69\\
SimMMDG~\cite{dong2023SimMMDG} & $\checkmark$& & $\checkmark$& \textbf{59.31} & 63.33 & 62.73 & 61.79&&79.31 & \textbf{77.04} & 51.29 & 69.21 \\
MOOSA (ours)& $\checkmark$&&  $\checkmark$  & 	 57.47	 & \textbf{65.47} & \textbf{63.66} &\textbf{62.20} & & 	 \textbf{80.39}	 & 76.27 &  \textbf{52.57}&  \textbf{69.74}\\

\midrule

DeepAll & & $\checkmark$&$\checkmark$ & 45.28 & 56.40 & 57.08 & 52.92& & 50.04 &  59.71 &38.97 &  49.57 \\
MM-SADA~\cite{Munro_2020_CVPR} & &$\checkmark$ &$\checkmark$ & 47.36  & 53.47  & 60.27   & 53.70 & & 46.58 & 61.81  &39.15 &  49.18 \\
RNA-Net~\cite{Planamente_2022_WACV} & & $\checkmark$&$\checkmark$ & 45.74 & 57.73 & 56.47 & 53.31&  & 52.05 & 64.13  & 40.35&  52.18\\
SimMMDG~\cite{dong2023SimMMDG} & & $\checkmark$&$\checkmark$ & 56.09 & 67.33 & 61.50 & 61.64 & &\textbf{59.63} & 64.24 & 44.85 & 56.24\\
MOOSA (ours)& & $\checkmark$& $\checkmark$  & 	\textbf{56.55}	 & \textbf{69.73} &  \textbf{62.53} & \textbf{62.94}  && 	59.41	 &  \textbf{65.01} & \textbf{45.31} & \textbf{56.58} \\

\midrule
DeepAll & $\checkmark$& $\checkmark$& $\checkmark$ & 55.63 & 59.20 & 58.01 & 57.61& &69.07&  71.30 &51.47 & 63.95  \\
MM-SADA~\cite{Munro_2020_CVPR} & $\checkmark$&$\checkmark$ &$\checkmark$ & 51.72  &  58.40 &  59.34  &   56.49&& 72.53 &  72.19 & 55.51&66.74   \\
RNA-Net~\cite{Planamente_2022_WACV} & $\checkmark$& $\checkmark$& $\checkmark$ & 52.41 & 57.20 & 60.16 & 56.59 &&  69.00&  73.40 & 51.65& 64.68 \\
SimMMDG~\cite{dong2023SimMMDG} & $\checkmark$& $\checkmark$& $\checkmark$ & \textbf{63.68} & 70.13 & 67.76 & 67.19& & 77.65 & \textbf{79.03} &56.62 & 71.10 \\
MOOSA (ours)& $\checkmark$& $\checkmark$& $\checkmark$  & 	63.22	 &  \textbf{70.27}& \textbf{68.38} & \textbf{67.29} & &	\textbf{78.01}	 & 77.15 &  \textbf{61.40} & \textbf{72.19} \\
\bottomrule
\end{tabular}

\end{threeparttable}
}
\vspace{0.1cm}
\caption{Multimodal \textbf{Closed-set DG} with different combinations of modalities on EPIC-Kitchens and HAC datasets.}
\label{tab:epic-dg2}
\end{table*}

\section{Ablation Study}

\begin{figure}[ht!]
  \centering  \includegraphics[width=\linewidth]{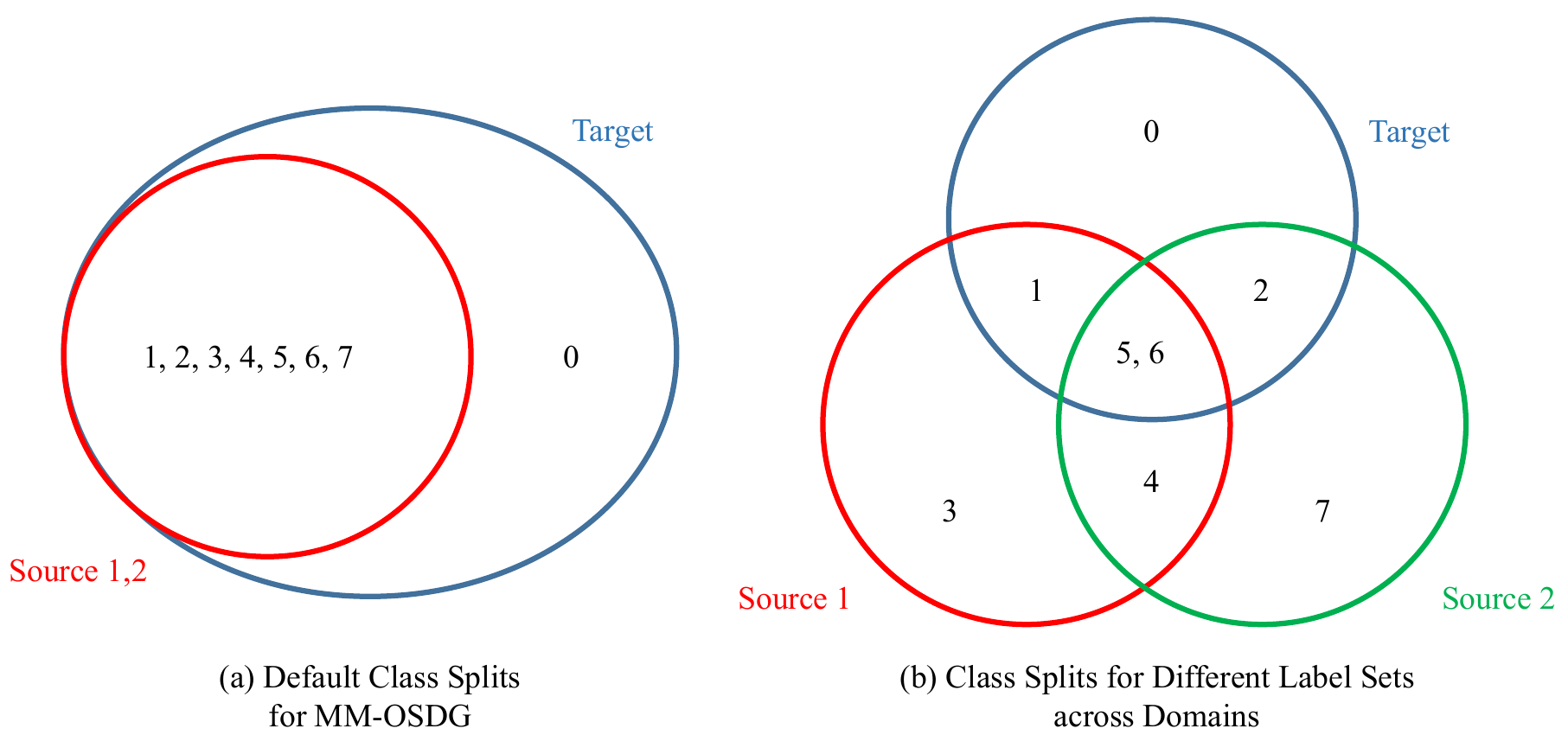}
   \caption{Class splits for different label sets across domains on EPIC-Kitchens dataset.}
   \label{fig:splits}
\end{figure}

\noindent\textbf{{Class Splits for Different Label Sets across Domains.}} In this setup, we assume different source domains have disparate label sets and the target domain does not necessarily include all label sets in source domains. This setup is more challenging and the details of label splits under this setup are shown in \cref{tab:diff-s} and \cref{fig:splits}.

\begin{table}[ht!]
\centering
\resizebox{0.32\textwidth}{!}{
\begin{tabular}{ccccc}
\hline Domain  & & && Classes \\
\hline Source-1  & & && $1,3,4,5,6$ \\
Source-2  & & && $2,4,5,6,7$ \\
Target & & & & $0,1,2,5,6$ \\
\hline
\end{tabular} 
}
\vspace{0.1cm}
\caption{Class splits for different label sets across domains on EPIC-Kitchens dataset.}
\label{tab:diff-s} 
\end{table}

\noindent\textbf{{Detailed Results on Different Openness.}}
\cref{tab:epic-openness1} to \cref{tab:epic-openness3} show the detailed results on different openness  (6:2, 5:3, 4:4). For the ratio of 6:2, we designate the first two classes in alphabetic order as unknown and the remaining classes as known, and we do the similar for 5:3 and 4:4. Our method consistently outperforms alternative approaches, showcasing significant improvements and validating the robustness of our model across diverse scenarios.

\begin{table}[t!]
\centering
\resizebox{\textwidth}{!}{
\begin{tabular}{l@{~~~~~}| c ccc ccc ccc ccc}
\hline

 \multicolumn{1}{c|}{{} } &\multicolumn{3}{c|}{D2, D3 $\rightarrow$ D1 } & \multicolumn{3}{c|}{D1, D3 $\rightarrow$ D2 } & \multicolumn{3}{c|}{D1, D2 $\rightarrow$ D3 }  &  \multicolumn{3}{c}{\textit{Mean} }
 \\
   \multicolumn{1}{c|}{{}} & OS* & UNK & \multicolumn{1}{c|}{\textbf{\underline{HOS}}} &  OS* & UNK &  \multicolumn{1}{c|}{\textbf{\underline{HOS}}}  & OS* & UNK &  \multicolumn{1}{c|}{\textbf{\underline{HOS}}}  & OS* & UNK &  \multicolumn{1}{c}{\textbf{\underline{HOS}}} \\
\hline

 \multicolumn{1}{c|}{DeepAll} & 39.43& 51.06 & \multicolumn{1}{c|}{44.50}  & 48.38 & 54.29 & \multicolumn{1}{c|}{51.16}  & 48.51 & 48.00 & \multicolumn{1}{c|}{48.25}  & 45.44 & 51.12 & \multicolumn{1}{c}{47.97}  \\

 \multicolumn{1}{c|}{{MM-SADA~\cite{Munro_2020_CVPR}} } &33.51 &  46.81& \multicolumn{1}{c|}{39.06}  &  44.85& 60.00 & \multicolumn{1}{c|}{51.33}  &46.57  & 48.00 & \multicolumn{1}{c|}{47.27}  & 41.64 & 51.60 & \multicolumn{1}{c}{45.89}  \\
 
 \multicolumn{1}{c|}{{RNA-Net~\cite{Planamente_2022_WACV}}} & 42.01&  57.45& \multicolumn{1}{c|}{48.53}  & 43.53 & 60.00 & \multicolumn{1}{c|}{50.45}  & 48.51 & 53.00 & \multicolumn{1}{c|}{50.66}  & 44.68 & 56.82 & \multicolumn{1}{c}{49.88}  \\

 \multicolumn{1}{c|}{{MEDIC~\cite{wang2023generalizable}}} & 36.86& 55.32 & \multicolumn{1}{c|}{44.24}  & 39.56 & 64.29 & \multicolumn{1}{c|}{48.98}  & 39.93& 57.00 & \multicolumn{1}{c|}{46.96}   & 38.78 & 58.87 & \multicolumn{1}{c}{46.73}  \\
 
 \multicolumn{1}{c|}{{SimMMDG~\cite{dong2023SimMMDG}} } &43.04 &  61.70& \multicolumn{1}{c|}{50.71}  & 43.82 & 62.86 & \multicolumn{1}{c|}{51.64}  & 45.08 & 66.00 & \multicolumn{1}{c|}{53.57}  & 43.98 & 63.52 & \multicolumn{1}{c}{51.97}  \\

 \multicolumn{1}{c|}{{MOOSA (ours)}} &45.36 & 59.57 & \multicolumn{1}{c|}{\textbf{51.51}}  &46.76  & 64.29 & \multicolumn{1}{c|}{\textbf{54.14}}  &48.40  & 69.00 & \multicolumn{1}{c|}{\textbf{56.89}}  & 46.84 & 64.29 & \multicolumn{1}{c}{\textbf{54.18}}  \\

\hline

\end{tabular} 
}
\vspace{0.1cm}
\caption{Abation on different openness (6:2) for Multimodal \textbf{Open-set} DG with video and audio modalities on EPIC-Kitchens dataset.}
\label{tab:epic-openness1} 
\end{table}

\begin{table}[t!]
\centering
\resizebox{\textwidth}{!}{
\begin{tabular}{l@{~~~~~}| c ccc ccc ccc ccc}
\hline

 \multicolumn{1}{c|}{{} } &\multicolumn{3}{c|}{D2, D3 $\rightarrow$ D1 } & \multicolumn{3}{c|}{D1, D3 $\rightarrow$ D2 } & \multicolumn{3}{c|}{D1, D2 $\rightarrow$ D3 }  &  \multicolumn{3}{c}{\textit{Mean} }
 \\
   \multicolumn{1}{c|}{{}} & OS* & UNK & \multicolumn{1}{c|}{\textbf{\underline{HOS}}} &  OS* & UNK &  \multicolumn{1}{c|}{\textbf{\underline{HOS}}}  & OS* & UNK &  \multicolumn{1}{c|}{\textbf{\underline{HOS}}}  & OS* & UNK &  \multicolumn{1}{c}{\textbf{\underline{HOS}}} \\
\hline

 \multicolumn{1}{c|}{DeepAll} & 37.77& 47.76 & \multicolumn{1}{c|}{42.18}  &48.17  &46.88  & \multicolumn{1}{c|}{47.51}  &  46.96& 46.53 & \multicolumn{1}{c|}{46.75}  & 44.30 & 47.06 & \multicolumn{1}{c}{45.48}  \\

 \multicolumn{1}{c|}{{MM-SADA~\cite{Munro_2020_CVPR}} } & 38.04&  49.25& \multicolumn{1}{c|}{42.93}  & 46.48 & 55.21 & \multicolumn{1}{c|}{50.47}  & 45.48 & 52.48 & \multicolumn{1}{c|}{48.73}  & 43.33 & 52.31 & \multicolumn{1}{c}{47.38}  \\
 
 \multicolumn{1}{c|}{{RNA-Net~\cite{Planamente_2022_WACV}}} & 42.66& 46.27 & \multicolumn{1}{c|}{44.39}  &49.85  & 55.21 & \multicolumn{1}{c|}{52.39}  & 45.13 & 37.62 & \multicolumn{1}{c|}{41.04}  & 45.88 & 46.37 & \multicolumn{1}{c}{45.94}  \\

 \multicolumn{1}{c|}{{MEDIC~\cite{wang2023generalizable}}} & 40.76& 61.19 & \multicolumn{1}{c|}{48.93}  & 38.84 &  69.79& \multicolumn{1}{c|}{49.90}  & 43.76& 57.43 & \multicolumn{1}{c|}{49.67}   &41.12  &  62.80& \multicolumn{1}{c}{49.50}  \\

 \multicolumn{1}{c|}{{SimMMDG~\cite{dong2023SimMMDG}} } &45.11 & 64.18 & \multicolumn{1}{c|}{52.98}  & 43.88 & 70.83 & \multicolumn{1}{c|}{54.19}  &42.50  &68.32  & \multicolumn{1}{c|}{52.40}  & 43.83 & 67.78 & \multicolumn{1}{c}{53.19}  \\

 \multicolumn{1}{c|}{{MOOSA (ours)}} &47.28 & 68.66 & \multicolumn{1}{c|}{\textbf{56.00}}  &  43.73& 73.96 & \multicolumn{1}{c|}{\textbf{54.96}}  & 48.45 & 65.35 & \multicolumn{1}{c|}{\textbf{55.65}}  & 46.49 & 69.32 & \multicolumn{1}{c}{\textbf{55.54}}  \\

\hline

\end{tabular} 
}
\vspace{0.1cm}
\caption{Abation on different openness (5:3) for Multimodal \textbf{Open-set} DG with video and audio modalities on EPIC-Kitchens dataset.}
\label{tab:epic-openness2} 
\end{table}

\begin{table}[t!]
\centering
\resizebox{\textwidth}{!}{
\begin{tabular}{l@{~~~~~}| c ccc ccc ccc ccc}
\hline

 \multicolumn{1}{c|}{{} } &\multicolumn{3}{c|}{D2, D3 $\rightarrow$ D1 } & \multicolumn{3}{c|}{D1, D3 $\rightarrow$ D2 } & \multicolumn{3}{c|}{D1, D2 $\rightarrow$ D3 }  &  \multicolumn{3}{c}{\textit{Mean} }
 \\
   \multicolumn{1}{c|}{{}} & OS* & UNK & \multicolumn{1}{c|}{\textbf{\underline{HOS}}} &  OS* & UNK &  \multicolumn{1}{c|}{\textbf{\underline{HOS}}}  & OS* & UNK &  \multicolumn{1}{c|}{\textbf{\underline{HOS}}}  & OS* & UNK &  \multicolumn{1}{c}{\textbf{\underline{HOS}}} \\
\hline

 \multicolumn{1}{c|}{DeepAll} & 47.15& 44.54 & \multicolumn{1}{c|}{45.81}  & 59.56 & 36.59 & \multicolumn{1}{c|}{45.33}  & 46.04 & 44.83 & \multicolumn{1}{c|}{45.43}  &  50.92& 41.99 & \multicolumn{1}{c}{45.52}  \\

 \multicolumn{1}{c|}{{MM-SADA~\cite{Munro_2020_CVPR}} } & 43.04& 36.13 & \multicolumn{1}{c|}{39.29}  & 54.95 & 39.63 & \multicolumn{1}{c|}{46.05}  &47.99  & 36.95 & \multicolumn{1}{c|}{41.75}  & 48.66 & 37.57 & \multicolumn{1}{c}{42.36}  \\
 
 \multicolumn{1}{c|}{{RNA-Net~\cite{Planamente_2022_WACV}}} & 48.42&  57.98& \multicolumn{1}{c|}{52.77}  &  59.56&  42.68& \multicolumn{1}{c|}{49.73}  & 51.75 & 35.47 & \multicolumn{1}{c|}{42.09}  & 53.24 & 45.38 & \multicolumn{1}{c}{48.20}  \\

 \multicolumn{1}{c|}{{MEDIC~\cite{wang2023generalizable}}} &48.42 & 51.26 & \multicolumn{1}{c|}{49.80}  & 51.71 & 48.78 & \multicolumn{1}{c|}{50.20}  & 41.76& 51.72 & \multicolumn{1}{c|}{46.21}   & 47.30 & 50.59 & \multicolumn{1}{c}{48.74}  \\
 
 \multicolumn{1}{c|}{{SimMMDG~\cite{dong2023SimMMDG}} } & 44.62&  65.55& \multicolumn{1}{c|}{53.10}  & 43.86 & 69.51 & \multicolumn{1}{c|}{53.78}  &  39.69&  53.69& \multicolumn{1}{c|}{45.64}  &42.72  & 62.92 & \multicolumn{1}{c}{50.84}  \\

 \multicolumn{1}{c|}{{MOOSA (ours)}} &52.53 & 59.66 & \multicolumn{1}{c|}{\textbf{55.87}}  &47.27  &65.24  & \multicolumn{1}{c|}{\textbf{54.82}}  &  40.34&  68.47& \multicolumn{1}{c|}{\textbf{50.77}}  & 46.71 & 64.46 & \multicolumn{1}{c}{\textbf{53.82}}  \\

\hline

\end{tabular} 
}
\vspace{0.1cm}
\caption{Abation on different openness (4:4) for Multimodal \textbf{Open-set} DG with video and audio modalities on EPIC-Kitchens dataset.}
\label{tab:epic-openness3} 
\end{table}

\noindent\textbf{{Different Confidence Score Functions for Unknown Class Detection.}} In our previous open-set experiments, we employed Maximum Softmax Probabilities (MSP) as the prediction confidence scores for detecting unknown classes. In addition to MSP~\cite{baseline_ood}, various metrics can serve as confidence scores, including Entropy~\cite{wang2020tent}, MaxLogit~\cite{hendrycks2019anomalyseg}, Energy~\cite{energy}, and Mahalanobis~\cite{mahalanobis}. As demonstrated in \cref{tab:epic-score}, despite its simplicity, MSP consistently delivers robust performances in most cases.  Conversely, while Energy and Mahalanobis are effective choices for unimodal OOD detection, they exhibit limitations in multimodal scenarios.

\begin{table}[t!]
\centering
\resizebox{\textwidth}{!}{
\begin{tabular}{l@{~~~~~}| c ccc ccc ccc ccc}
\hline

 \multicolumn{1}{c|}{{Self-supervised} } &\multicolumn{3}{c|}{D2, D3 $\rightarrow$ D1 } & \multicolumn{3}{c|}{D1, D3 $\rightarrow$ D2 } & \multicolumn{3}{c|}{D1, D2 $\rightarrow$ D3 }  &  \multicolumn{3}{c}{\textit{Mean} }
 \\
   \multicolumn{1}{c|}{{Tasks}} & OS* & UNK & \multicolumn{1}{c|}{\textbf{\underline{HOS}}} &  OS* & UNK &  \multicolumn{1}{c|}{\textbf{\underline{HOS}}}  & OS* & UNK &  \multicolumn{1}{c|}{\textbf{\underline{HOS}}}  & OS* & UNK &  \multicolumn{1}{c}{\textbf{\underline{HOS}}} \\
\hline

 \multicolumn{1}{c|}{MSP~\cite{baseline_ood}} &  41.90 &82.35& \multicolumn{1}{c|}{55.54}  & 45.53 & 64.29 & \multicolumn{1}{c|}{53.31} & 45.52 & 69.01& \multicolumn{1}{c|}{54.85}  &44.32  & 71.88 & \multicolumn{1}{c}{54.57} \\
 
 \multicolumn{1}{c|}{Entropy~\cite{wang2020tent}} & 42.14 &73.53  & \multicolumn{1}{c|}{53.58}  & 42.94 & 75.00 & \multicolumn{1}{c|}{54.61}  & 46.18 &  64.79& \multicolumn{1}{c|}{53.92}  &  43.75& 71.11 & \multicolumn{1}{c}{54.04}  \\
 
 \multicolumn{1}{c|}{MaxLogit~\cite{hendrycks2019anomalyseg}} & 42.89 & 52.94 & \multicolumn{1}{c|}{47.39}  & 33.00 & 57.14 & \multicolumn{1}{c|}{41.84}  & 30.56 & 63.38 & \multicolumn{1}{c|}{41.24}  & 35.48 & 57.82 & \multicolumn{1}{c}{43.49}  \\

 \multicolumn{1}{c|}{Energy~\cite{energy}} & 37.41 & 61.76 & \multicolumn{1}{c|}{46.59}  & 42.36 & 41.07 & \multicolumn{1}{c|}{41.71}  & 43.96 & 38.03 & \multicolumn{1}{c|}{40.78}  & 41.24 & 46.95 & \multicolumn{1}{c}{43.03}  \\
 
  \multicolumn{1}{c|}{Mahalanobis~\cite{mahalanobis}}& 17.71 & 70.59 & \multicolumn{1}{c|}{28.31}  & 38.04 & 57.14 & \multicolumn{1}{c|}{45.67}  & 40.53 & 57.75 & \multicolumn{1}{c|}{47.63}  & 32.09 & 61.83 & \multicolumn{1}{c}{40.54}  \\

\hline

\end{tabular} 
}
\vspace{0.1cm}
\caption{Abation on different OOD score functions as confidence scores for MM-OSDG with video and audio modalities on EPIC-Kitchens dataset.}
\label{tab:epic-score} 
\end{table}

\begin{figure}[t!]
  \centering  \includegraphics[width=0.9\linewidth]{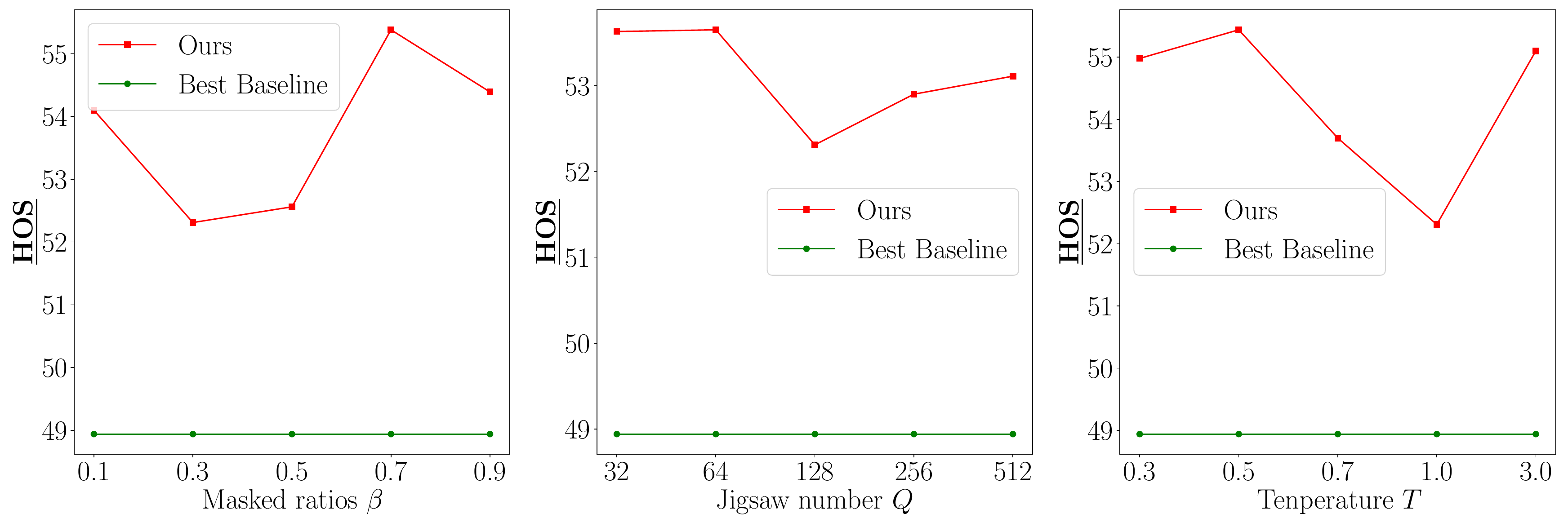}
   \caption{Parameter sensitivity to the hyperparameters in the self-supervised tasks.}
   \label{fig:sensitivity}
\end{figure}

\begin{figure}[t!]
  \centering  \includegraphics[width=0.9\linewidth]{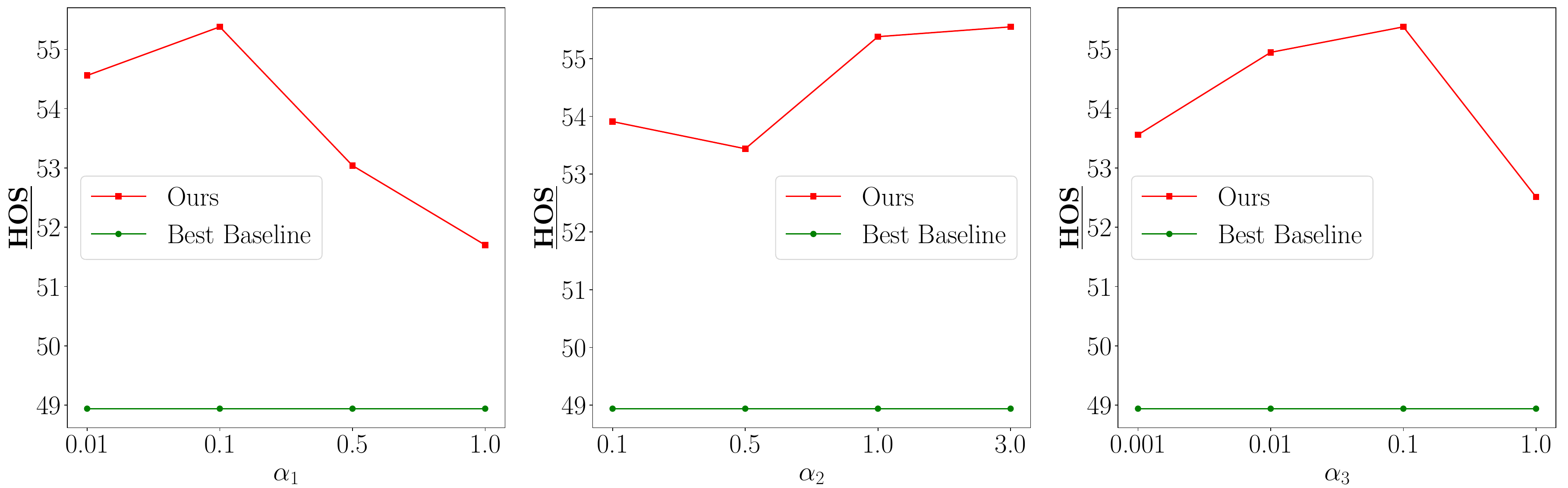}
   \caption{Parameter sensitivity to the hyperparameters in the loss.}
   \label{fig:losssensitivity}
\end{figure}

\noindent\textbf{{Parameter Sensitivity Analyze.}} 
We investigate the sensitivity of our method to the hyperparameters in the self-supervised tasks and loss function. We perform comprehensive analysis by varying one parameter while keeping others fixed, and the results are illustrated in~\cref{fig:sensitivity} and \cref{fig:losssensitivity}.
Remarkably, our method consistently surpasses the best baseline across all parameter settings, suggesting that our approach exhibits less sensitivity to hyperparameter choices.

\begin{figure}[h!]
  \centering  \includegraphics[width=0.9\linewidth]{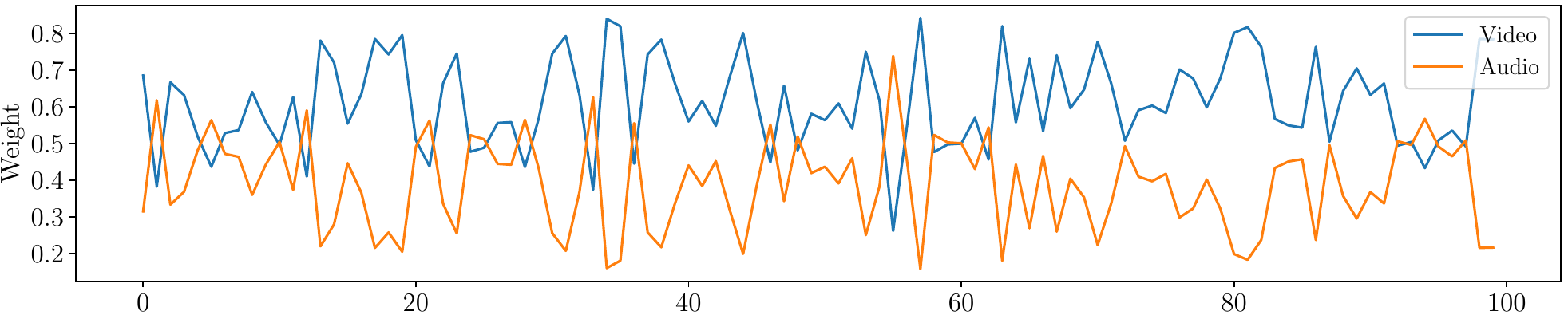}
   \caption{Examples of weights in Entropy Weighting for video and audio.}
   \label{fig:ent}
\end{figure}

\noindent\textbf{The Weight Distribution across Modalities in Entropy Weighting.} When training using video and audio on EPIC-Kitchens, the average weight for video is $0.63$ and $0.37$ for audio, indicating that video carries more useful information for prediction. \cref{fig:ent} shows examples of weights from random $100$ samples.

\section{Motivation behind MOOSA for Open-set DG Setting}

For Open-set DG setting, we need to address both \textbf{domain shift} and \textbf{category shift}. Self-supervised tasks (i.e., solving Jigsaw puzzles~\cite{carlucci2019domain}, predicting image rotations~\cite{bucci2020effectiveness}) have shown their efficacy in mitigating \textbf{domain shift} by capturing invariance to bridge domain gaps. Inspired by solving unimodal Jigsaw puzzles on images, we propose the novel \textit{Multimodal Jigsaw Puzzles (MulJig)} in multimodal scenarios to learn domain-invariant features and address the high heterogeneity across modalities by random shuffling in the embedding space. For \textbf{category shift}, the key lies in detecting unknown classes in target domain. Masked image modeling~\cite{li2023rethinking} has shown its efficacy in unknown class detection by forcing the network to learn the real data distribution of known class samples during training, enhancing the divergence between known and unknown samples. Inspired by this, we introduce the novel \textit{Masked Cross-modal Translation (MaskedTrans)} to facilitate the model in learning intrinsic data distributions for known classes across diverse modalities. \textit{MaskedTrans} and \textit{MulJig} are inherently \textbf{complementary} in addressing both domain and category shifts effectively.

\end{document}